\documentclass[]{fairmeta}
\PassOptionsToPackage{table}{xcolor}

\input{tmi.sty}

\usepackage[T1]{fontenc}
\usepackage{etoolbox}
\usepackage{amssymb,amsfonts,amsmath,mathtools}
\usepackage{graphicx}
\usepackage{textcomp}
\usepackage{array}
\usepackage{float}
\usepackage{xparse}
\usepackage{makecell}
\usepackage{pifont}
\usepackage{placeins}
\usepackage{dblfloatfix}
\usepackage{multicol}
\usepackage{xspace}
\usepackage{multirow}
\usepackage{tabularx}
\usepackage{booktabs}
\usepackage{algorithm}
\usepackage{algpseudocode}
\usepackage{siunitx}
\usepackage{url}
\usepackage{enumitem}
\usepackage{wrapfig}
\usepackage{caption}
\usepackage{etoolbox}
\usepackage{nicematrix}
\usepackage{cleveref}
\usepackage{lipsum}
\usepackage{fontawesome5}
\usepackage{scalefnt}
\usepackage{subcaption}
\usepackage{setspace}
\usepackage{soul}
\usepackage{calc}
\usepackage{hhline}
\usepackage{diagbox}
\usepackage{extarrows}
\usepackage{fancyvrb}
\usepackage{adjustbox}
\usepackage{floatflt}
\usepackage{bm}
\usepackage{microtype}
\usepackage{wasysym}
\usepackage{titlecaps}
\setcitestyle{numbers,square,comma,sort,compress}
\graphicspath{{./}}

\newcommand{\bench}{\textsc{MedGEN-Bench}\xspace}
\newcommand{\medgenciterange}[2]{\nocite{#1,#2}\mbox{[\csname b@#1\endcsname\textendash\csname b@#2\endcsname]}}
\newcommand{\unifiedmodel}{\includegraphics[height=1em]{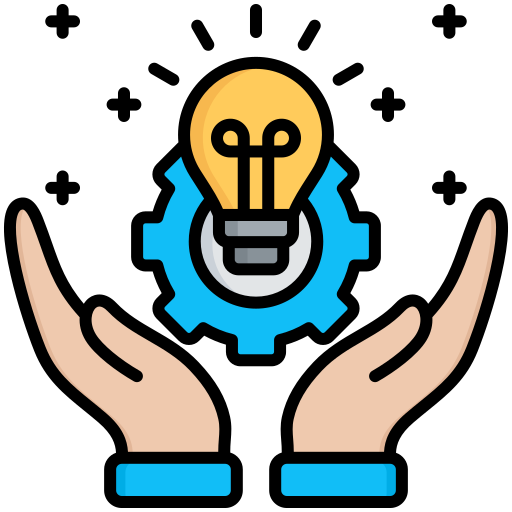}\xspace}
\newcommand{\compositional}{\includegraphics[height=1em]{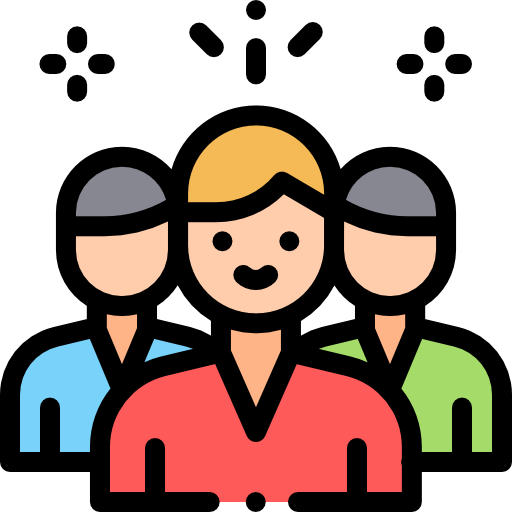}\xspace}
\makeatletter
\newcommand{\medgencaptionof}[1]{\def\@captype{#1}\caption}
\makeatother

\makeatletter
\@ifundefined{MarkedUpCopy}{\newcommand{\MarkedUpCopy}{0}}{}
\makeatother
\newcommand{\revisioncolor}{red}
\ifnum\MarkedUpCopy=1
  \newcommand{\revisionon}{\color{\revisioncolor}}
  \newcommand{\revnote}[1]{{\revisionon [Revision note: #1]}}
\else
  \newcommand{\revisionon}{\color{black}}
  \newcommand{\revnote}[1]{}
\fi
\newcommand{\revisionoff}{\color{black}}
\newcommand{\rev}[1]{{\revisionon #1}}

\newcommand{\marknewreference}[1]{
  \ifstrequal{#1}{lingshu2025}{\revisionon}{%
  \ifstrequal{#1}{zhang2025llmevalmed}{\revisionon}{%
  \ifstrequal{#1}{ding2025medrewardbench}{\revisionon}{%
  \ifstrequal{#1}{cooke2026roentmod}{\revisionon}{%
  \ifstrequal{#1}{delbrouck2022radgraph}{\revisionon}{%
  \ifstrequal{#1}{green}{\revisionon}{%
  \ifstrequal{#1}{radcliq}{\revisionon}{}}}}}}}%
}

\newcommand{\IEEEPARstart}[2]{#1#2}

\newcommand{\IEEEtriggeratref}[1]{}
\renewcommand{\journalname}{arXiv migration build}
\makeatletter
\let\medgen@origabstract\abstract
\long\def\abstract#1{\gdef\abstractlist{{\color{metafg} #1}}}
\makeatother
\title{MedGEN-Bench: A Contextually Entangled Benchmark for Open-Ended Multimodal \\ Medical Generation}
\author[1]{Junjie Yang}
\author[2]{Yuhao Yan}
\author[3]{Gang Wu}
\author[5]{Rui Qian}
\author[6]{Zhisheng Chen}
\author[9]{Haijiang Li}
\author[12]{Yuhe Wu}
\author[13]{Qichao Zhao}
\author[11]{Dawen Tian}
\author[4]{Xiang Wan}
\author[10]{Fenglei Fan}
\author[7]{Wenjian Qin}
\author[8]{Yongquan Zhang}
\author[3]{Feiwei Qin}
\author[4]{Changmiao Wang}

\affiliation[1]{South China University of Technology}
\affiliation[2]{Sun Yat-sen University}
\affiliation[3]{Hangzhou Dianzi University}
\affiliation[4]{Shenzhen Research Institute of Big Data}
\affiliation[5]{Fudan University}
\affiliation[6]{Nanyang Technological University}
\affiliation[7]{Shenzhen Institute of Advanced Technology}
\affiliation[8]{Zhejiang University of Finance and Economics}
\affiliation[9]{Independent}
\affiliation[10]{The Hong Kong Polytechnic University}
\affiliation[11]{Xiangtan University}
\affiliation[12]{The Hong Kong University of Science and Technology}
\affiliation[13]{Tsinghua University}

\metadata[Home page]{\url{https://yangjj007.github.io/medgen}}
\metadata[Dataset]{\url{https://huggingface.co/datasets/Jack04810/MedGEN-Bench}}
\metadata[Code]{\url{https://github.com/yangjj007/MedGEN-Bench-eval}}
\metadata[Mail]{\texttt{\{yangjj.jack, yanyuhao, qiianruii, cmwangalbert\}@gmail.com}}

\hypersetup{hidelinks}
\vspace{-2cm}
\begin{document}
\abstract{
\rev{Medical vision-language models (VLMs) are increasingly expected to support clinical workflows through the generation of diagnostic text and relevant medical images. However, existing medical visual benchmarks exhibit three recurring limitations: query--image misalignment arising from queries that are weakly grounded in specific image instances, closed-ended formats that narrow the answer space and encourage shortcut-based prediction, and text-centric output paradigms that limit the evaluation of image-generation and image-editing capabilities. To address these limitations, we introduce \bench, a benchmark for open-ended multimodal medical generation. The evaluation snapshot reported in this manuscript comprises 6,422 image--text pairs reviewed by clinical experts and models, spanning 6 canonical imaging modalities, 15 clinical tasks, and 27 named subtasks. It includes 1,100 Visual Question Answering (VQA) pairs, 3,872 Image Editing pairs, and 1,450 Contextual Multimodal Generation pairs. \bench centers on contextual entanglement, defined as the dependence of an instruction's intended output on the particular image instance rather than on task wording alone. The benchmark operationalizes this concept through image-grounded instructions and extends evaluation to open-ended multimodal outputs. Its tiered evaluation protocol combines reproducible, reference-based fidelity and similarity measures with a structured, checklist-guided assessment by a medical VLM judge. We evaluate 10 compositional frameworks, 2 dedicated image-editing models, 3 unified models, and 5 VLMs. The results show that image-output tasks remain unsaturated. Contextual augmentation increases the mean image--instruction similarity from 0.273 to 0.372, while a 1,000-case medical-expert audit shows moderate agreement between the judge scores and clinician ratings. The source code and dataset are available on \href{https://yangjj007.github.io/medgen}{\textcolor{red}{GitHub}}.}}


\maketitle

\vspace{-0.03cm}
\section{Introduction}
\label{sec:intro}

\begin{figure*}[htb]
\centering
\vspace{-1.5em}
\includegraphics[width=1.0\linewidth]{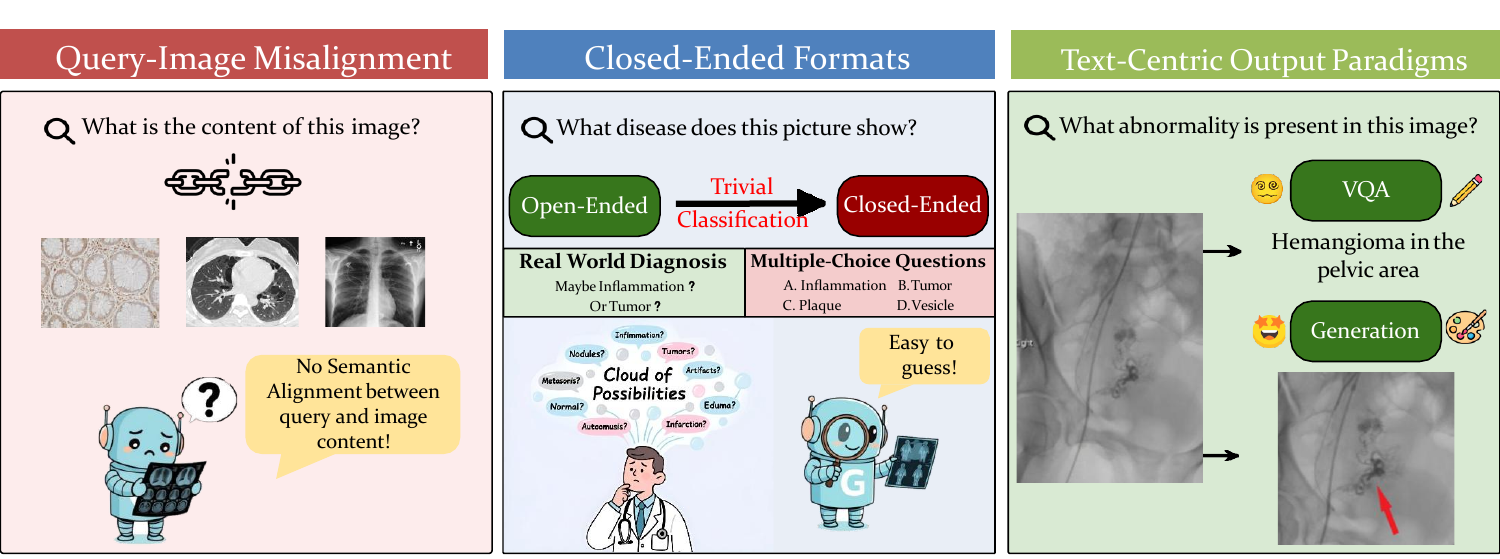}
\caption{\rev{Our pilot study highlights three recurring limitations of existing medical visual benchmarks: (1) query--image misalignment arising when generic or templated queries are weakly grounded in specific image instances, (2) closed-ended formats that narrow the answer space and encourage shortcut reliance, and (3) text-centric output paradigms that limit evaluation of image-generation and image-editing capabilities.}}
\label{fig:pilot_study}
\end{figure*}


\IEEEPARstart{T}{he} \rev{growing adoption of vision-language models (VLMs) in medicine creates demand for systems capable of generating diagnostic text and relevant medical image outputs \cite{ye2024gmai,alayrac2022flamingo}. However, current medical VLM evaluation primarily emphasizes Visual Question Answering (VQA) tasks with text-only outputs \cite{zhang2023pmc}. This emphasis does not fully reflect clinical workflows, in which physicians integrate patient history, imaging, and longitudinal evidence for tasks such as lesion localization, prediction, and planning \cite{zhou2025drvd}. Evaluating support for such workflows requires testing whether models ground instructions in individual images and extend beyond text generation to medical image transformation. We therefore focus on contextual entanglement, operationally defined as the dependence of an intended answer or transformation on the particular image instance rather than on task wording alone. Evaluating this property directly tests image-specific task grounding. Despite advances in medical VQA and generative medical imaging \cite{MedGAN,SynthRAD}, comprehensive evaluation of open-ended multimodal generation remains limited.}

\rev{A pilot analysis of prominent medical visual benchmarks identifies three recurring limitations: (1) query--image misalignment that arises when generic or templated queries are weakly grounded in specific image instances, (2) closed-ended formats that narrow the answer space and encourage shortcut reliance, and (3) text-centric output paradigms that limit the evaluation of image-generation and image-editing capabilities. As illustrated in Figure~\ref{fig:pilot_study}, these limitations motivate the development of more comprehensive benchmarks that support image-grounded, open-ended multimodal evaluation.}


\color{black}
\rev{To this end, we introduce the \textbf{Med}ical \textbf{G}enerative \textbf{EN}tangled Benchmark (\bench), a multimodal evaluation framework for medical AI. We comprises 6,422 pairs reviewed by clinical experts and models, referencing 10,703 unique images. Each record links a medical image and an image-grounded instruction to a task-dependent target output in the form of text, an edited or generated image, or both. For reporting, the \bench uses 6 canonical modality labels: CT, MRI, Ultrasound, X-ray, Pathology, and Clinical Photography. Records whose source metadata cannot be mapped reliably are retained in an ``other/unspecified'' category in the manifest.}

\rev{Our \bench covers 15 clinical tasks and 27 named subtasks organized into 3 formats: Visual Question Answering with 1,100 pairs, Image Editing with 3,872 pairs, and Contextual Multimodal Generation with 1,450 pairs. The four VQA task formats have no additional named subtasks. As illustrated in Figure~\ref{fig:pipeline}, benchmark construction follows a four-stage pipeline comprising two-stage image filtering, rule-based and model-based image-pair synthesis, context-augmented text-pair synthesis, and postprocessing through automatic checks and expert review. The 6,422 released pairs were reviewed by clinical experts and models; Appendix~\ref{app:review} provides the detailed procedure. \bench uses image-grounded instructions and extends evaluation beyond closed-form VQA to image editing and open-ended multimodal outputs. It evaluates full-image transformations alongside text outputs.}

\rev{Using \bench, we establish baseline results for 10 compositional frameworks, 2 dedicated image-editing models, 3 unified models, and 5 VLMs, and explicitly report their output coverage. Reference-based fidelity and similarity measures are reported separately alongside raw holistic judge scores. The results show substantial headroom in image-output tasks. In compositional pipelines, point estimates vary across configurations in separate generation and editing blocks. Contextual augmentation increases mean image--instruction similarity from 0.273 to 0.372, corresponding to a relative increase of 36.3\%. Behavioral pairing-sensitivity analysis shows that disrupting the image--instruction pairing reduces cross-modal consistency for contextual multimodal-generation outputs. A 1,000-case medical-expert audit shows moderate agreement between the judge scores and clinician ratings. Together, these analyses position \bench as a benchmarking resource for diagnosing failures in image-specific grounding and open-ended multimodal generation.}
\color{black}
\rev{In summary, our work makes four key contributions:}
\begin{itemize}
    \item \rev{We identify three recurring limitations: query–image misalignment, closed-ended formats, and text-centric output paradigms, all of which restrict the evaluation of image grounding, open-ended answers, image generation, and image editing capabilities.}
    \item \rev{We introduce \bench: 6,422 pairs reviewed by clinical experts and models, referencing 10,703 unique images across 6 modalities, 15 clinical tasks, and 27 named subtasks.}
    \item \rev{We establish a three-tier framework combining reference-based image fidelity, text similarity, answer agreement, and checklist-guided assessment by medical VLM judge.}
    \item \rev{We evaluate 10 compositional frameworks, 2 image-editing models, 3 unified models, and 5 VLMs, providing baseline evidence on image-output performance, configuration sensitivity, and image-specific task grounding.}
\end{itemize}

\section{Related Work}
\label{sec:related}

\begin{table*}[!t]
\centering
\vspace*{-1.75em}
\tabcolsep=0.1cm
\caption{Comparison of medical visual benchmarks by imaging modality, output format, contextual entanglement, and open-ended answering.}
\label{tab:Comparison}
\resizebox{1.0\textwidth}{!}{
\begin{tabular}{@{}lllcc@{}}
\hline\hline
\textbf{Benchmark} & \textbf{Image Modalities} & \textbf{Format Types} & \textbf{Contextual Entanglement} & \textbf{Open-ended Answering} \\ \hline
VQA-RAD \cite{lau2018dataset} & X-Ray, CT, MRI & VQA & \textcolor{red}{\ding{55}} & \textcolor{green!60!black}{\checkmark} \\ 
SLAKE \cite{liu2021slake} & X-Ray, CT, MRI & VQA & \textcolor{red}{\ding{55}} & \textcolor{green!60!black}{\checkmark} \\ 
PMC-VQA \cite{zhang2023pmc} & X-Ray, CT, MRI and others & VQA & \textcolor{red}{\ding{55}} & \textcolor{green!60!black}{\checkmark} \\ 
PathVQA \cite{he2020pathvqa} & Pathology & VQA & \textcolor{red}{\ding{55}} & \textcolor{green!60!black}{\checkmark} \\ 
MIMIC-Diff-VQA \cite{hu2023diffvqa} & Chest X-ray & Difference VQA & \textcolor{green!60!black}{\checkmark} & \textcolor{green!60!black}{\checkmark} \\ 
Quilt-LLaVA / Quilt-VQA \cite{seyfioglu2024quiltllava} & Histopathology & Localized VQA & \textcolor{red}{\ding{55}} & \textcolor{green!60!black}{\checkmark} \\ 
ImageCLEF VQA-Med \cite{abacha2019vqamed,abacha2021vqamed} & Radiology and other medical images & VQA / answer generation & \textcolor{red}{\ding{55}} & \textcolor{green!60!black}{\checkmark} \\ 
OmniMedVQA \cite{omnimedvqa} & 12 modalities & VQA & \textcolor{red}{\ding{55}} & \textcolor{red}{\ding{55}} \\ 
GMAI-MMBench \cite{ye2024gmai} & 5 modalities & VQA & \textcolor{red}{\ding{55}} & \textcolor{red}{\ding{55}} \\ 
CARES \cite{xia2024cares} & 16 modalities & VQA & \textcolor{red}{\ding{55}} & \textcolor{green!60!black}{\checkmark} \\ 
MedFrameQA \cite{yu2025medframeqa} & X-Ray, CT, MRI & Multi-Image VQA & \textcolor{green!60!black}{\checkmark} & \textcolor{red}{\ding{55}} \\ 
DrVD-Bench \cite{zhou2025drvd} & 5 modalities & VQA, Report generation & \textcolor{red}{\ding{55}} & \textcolor{green!60!black}{\checkmark} \\ 
SMMILE \cite{smmile} & 16 modalities & Multimodal ICL (VQA-style)  & \textcolor{red}{\ding{55}} & \textcolor{green!60!black}{\checkmark} \\ 
MedJourney \cite{wu2024medjourney} & Clinical records / journeys & Longitudinal clinical benchmark & \textcolor{red}{\ding{55}} & \textcolor{green!60!black}{\checkmark} \\ 
CheXGenBench \cite{dutt2025chexgenbench} & X-Ray (Chest) & Multimodal Generation & \textcolor{red}{\ding{55}} & \textcolor{red}{\ding{55}} \\
MedEBench \cite{liu2025medebench} & 13 anatomical regions & Image Editing & \textcolor{red}{\ding{55}} & \textcolor{red}{\ding{55}} \\
\hline
\textbf{MedGEN-Bench} & 6 modalities & VQA, Image Editing, Multimodal Generation & \textcolor{green!60!black}{\checkmark} & \textcolor{green!60!black}{\checkmark} \\ 
\hline\hline
\end{tabular}
}
\end{table*}
\color{black}

\rev{\textbf{Medical VLMs and Generative Models.}}
\rev{General Large Vision-Language Models (LVLMs)~\cite{alayrac2022flamingo,li2023blip,achiam2023gpt} provide a strong foundation for general multimodal understanding and alignment, but medical applications impose additional requirements for domain knowledge, clinical image interpretation, and anatomical fidelity. Specialized models such as Med-Flamingo~\cite{moor2023med}, LLaVA-Med~\cite{li2023llava}, and the Med-PaLM series~\cite{singhal2023large} integrate medical images with clinical text and achieve strong performance on medical VQA and diagnostic tasks. In parallel, diffusion models~\cite{ho2020denoising,rombach2022high} have advanced image synthesis. Their medical applications~\cite{kazerouni2023diffusion}, however, typically focus on reconstruction-oriented tasks, including MRI and CT super-resolution, with less attention to high-level semantic generation and instruction-guided editing. Consequently, systematic evaluation of anatomically plausible and clinically meaningful medical image generation and editing remains limited.}

\setcounter{dbltopnumber}{1}
\AddToHookNext{shipout/after}{\AddToHookNext{shipout/after}{\AddToHookNext{shipout/after}{\setcounter{dbltopnumber}{4}}}}
\begin{figure*}[t]
\centering
\vspace{-1.8em}
\includegraphics[width=1.0\linewidth]{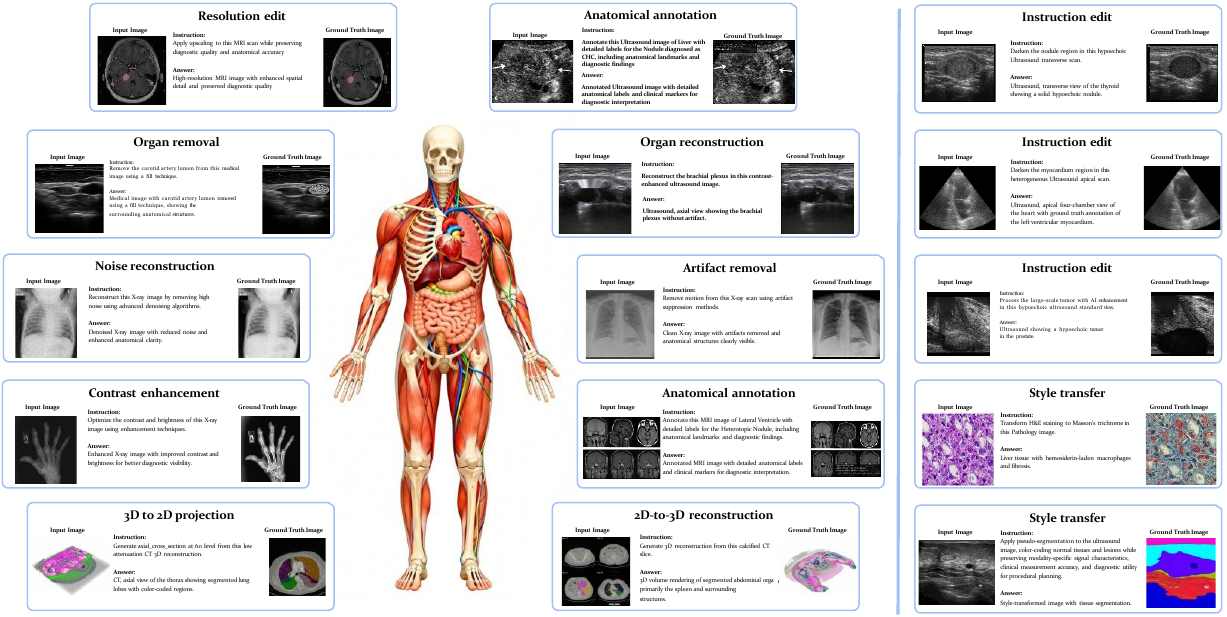}
\vspace{-1.5em}
\caption{\rev{Representative task cards in \bench. The examples cover image editing tasks involving resolution, noise, artifact, and contrast correction, style transfer, and anatomical annotation, as well as contextual multimodal generation tasks involving instruction-guided editing, organ-removal and organ-reconstruction, and 2D-to-3D reconstruction or 3D-to-2D projection. The central anatomy illustration summarizes coverage across body systems. Each card displays an input image, image-grounded instruction, and reviewed target image or answer.}}
\color{black}
\vspace{-0.2em}
\label{fig:examples}
\end{figure*}

\rev{\textbf{Evaluation Metrics.}}
\rev{Text generation by medical multimodal models is commonly evaluated using n-gram-based metrics such as BLEU~\cite{papineni2002bleu}, ROUGE~\cite{lin2004rouge}, and CIDEr~\cite{vedantam2015cider}. These metrics quantify lexical overlap but do not directly capture semantic equivalence or clinical factuality. Embedding-based metrics such as BERTScore~\cite{zhang2019bertscore} extend this evaluation by comparing contextual representations derived from BERT~\cite{devlin2019bert}, thereby providing a more informative measure of semantic similarity. For image-generation tasks, Inception Score (IS)~\cite{salimans2016improved} and Fréchet Inception Distance (FID)~\cite{heusel2017gans} measure general image quality or distributional similarity but do not directly assess anatomical fidelity or clinically significant abnormalities. As highlighted by~\cite{deo2025metrics}, general metrics can overestimate performance while missing clinical errors, thereby motivating specialized evaluation frameworks in medical contexts.}

\rev{\textbf{Medical Benchmarks.}}
\rev{Existing medical LVLM benchmarks primarily emphasize comprehension. VQA-RAD~\cite{lau2018dataset}, SLAKE~\cite{liu2021slake}, PMC-VQA~\cite{zhang2023pmc}, and ImageCLEF VQA-Med~\cite{abacha2019vqamed,abacha2021vqamed} assess medical VQA, while PathVQA~\cite{he2020pathvqa} and Quilt-LLaVA/Quilt-VQA~\cite{seyfioglu2024quiltllava} support pathology evaluation at different spatial scales. MIMIC-Diff-VQA~\cite{hu2023diffvqa} grounds questions in differences between paired chest radiographs, providing an image-conditioned comparison setting. MedJourney~\cite{wu2024medjourney} evaluates longitudinal patient journeys but not image generation. OmniMedVQA~\cite{omnimedvqa}, GMAI-MMBench~\cite{ye2024gmai}, and MedFrameQA~\cite{yu2025medframeqa} broaden clinical-reasoning coverage. DrVD-Bench~\cite{zhou2025drvd} evaluates reasoning consistency, whereas SMMILE~\cite{smmile} focuses on few-shot learning. Together, these benchmarks provide task-, modality-, or longitudinal evaluation. CheXGenBench~\cite{dutt2025chexgenbench} and MedEBench~\cite{liu2025medebench} cover image generation/editing but focus on specific modalities or task settings. Table~\ref{tab:Comparison} shows that \bench complements these benchmarks by combining 6 canonical imaging modalities, 15 clinical tasks, and 3 output formats in evaluation. Note that the checkmark for contextual entanglement denotes the benchmark's design objective rather than exhaustive validation of image grounding.}

\section{Limitations of Existing Medical VQA Benchmark: A Pilot Study}
\color{black}

\rev{This pilot study summarizes limitations of existing medical VQA benchmarks that motivate our benchmark design. It highlights the need for image-specific task grounding, open-ended evaluation, and support for image generation and editing.}

\rev{\textbf{Query--Image Misalignment.}}  
\rev{Generic or templated queries can be weakly grounded in the specific image instances they reference, creating query--image misalignment. This query design can reduce medical VQA to simplified classification or captioning exercises~\cite{agrawal2018dontjustassumelook}. Models can then use query keywords for task selection instead of robust visual-textual reasoning~\cite{agrawal2018dontjustassumelook,chen2020counterfactualsamplessynthesizingrobust}. This stands in stark contrast to clinical practice, where diagnostic queries are highly specific and intricately tied to particular visual findings, requiring image-specific grounding and contextual understanding~\cite{zhou2025drvd}.}

\rev{\textbf{Closed-Ended Formats.}}  
\rev{Closed-ended formats, such as multiple-choice questions, narrow the answer space and can encourage shortcut reliance. By limiting responses to predefined options, they introduce evaluation bias through information leakage \cite{agrawal2018dontjustassumelook} and transform complex tasks such as open-ended differential diagnosis into ranking exercises \cite{chen2020counterfactualsamplessynthesizingrobust}. This approach oversimplifies clinical reasoning~\cite{zhou2025drvd} and provides limited evidence about a model's capacity to generate open-ended diagnostic hypotheses.} \rev{The same CT input can yield a valid answer from choices but an unsupported report, as Figure~\ref{fig:closed_open_contrast} shows.}

\rev{\textbf{Text-Centric Output Paradigms.}}  
\rev{Text-centric output paradigms map image-text inputs to text-only outputs \cite{lau2018dataset,he2020pathvqa,liu2021slake}, limiting evaluation of image-generation and image-editing capabilities. Although recent resources extend evaluation to image-generation or image-editing settings, image-output capabilities and image-specific instruction grounding remain underexamined. In clinical settings, professionals perform tasks such as lesion localization, anatomical segmentation, and region annotation. These capabilities are not evaluated by benchmarks restricted to textual responses.}




\vspace*{-0.25em}
\section{Benchmark}
\vspace*{-0.25em}

\subsection{Overview}
\color{black}



\begin{figure*}[t]
\centering
\includegraphics[width=1.0\linewidth]{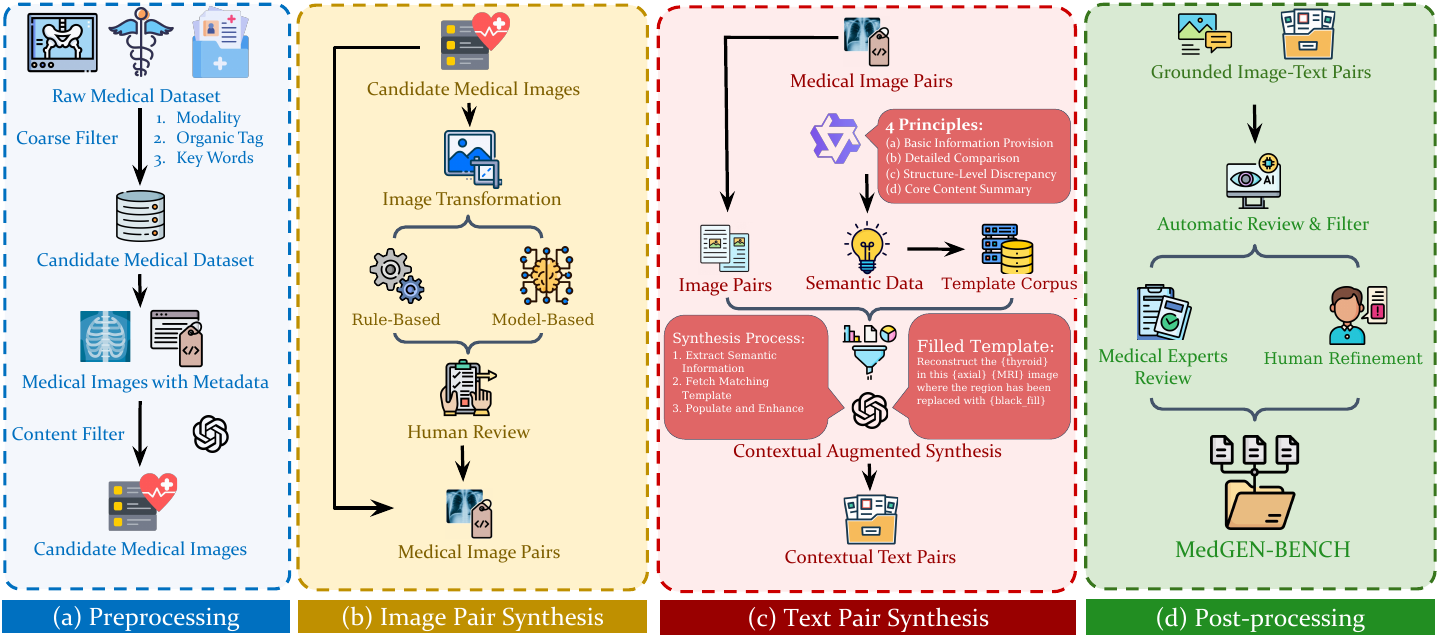}
\caption{{Overview of the \bench construction pipeline.} To show how raw medical data become benchmark pairs reviewed by clinical experts and models, the figure organizes the pipeline into four sequential phases. (\rev{a}) Preprocessing (blue) uses metadata-based coarse filtering followed by GPT-4o semantic validation to select candidate medical images and associated metadata. (\rev{b}) Image Pair Synthesis (yellow) combines deterministic operations with generative image transformations. (\rev{c}) Text Pair Synthesis (red) uses Qwen3-VL to extract semantic information and populate task-specific templates, after which GPT-4o integrates contextual queries with visual content. (\rev{d}) Postprocessing (green) uses automated Vision-Language Model review and review by clinical experts and GPT-4o to ensure high output quality.}
\label{fig:pipeline}
\end{figure*}

\rev{\bench is developed using a four-stage pipeline to create open-ended, clinically relevant tasks: (1) medical image preprocessing to ensure task relevance; (2) image-pair synthesis using rule-based and generative methods; (3) large-language-model-based text-pair synthesis for instruction--answer pairs; and (4) postprocessing with automated checks and expert review. Figure~\ref{fig:pipeline} summarizes the complete workflow and its quality-assurance procedures.}

\begin{figure*}[htb]
\centering
\includegraphics[width=1.0\textwidth]{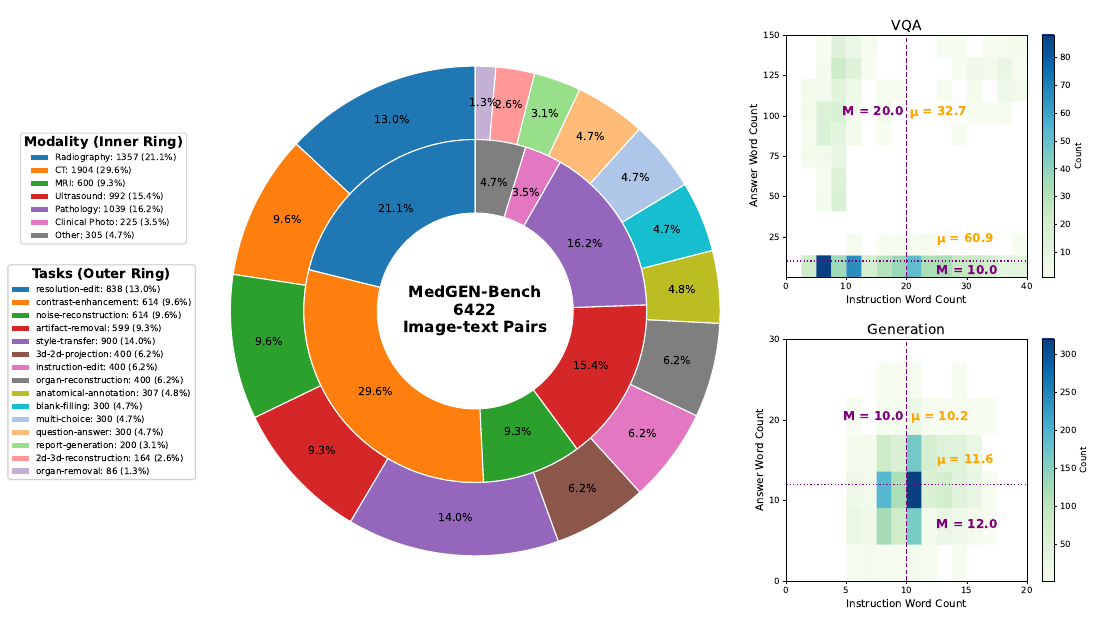}
\vspace{-1.1em}

\caption{\rev{\textbf{Left:} Overview of \bench statistics. \textbf{Right:} Instruction- and answer-length distributions measured by word count. \textbf{Key observation:} VQA lengths are strongly right-skewed, whereas generation text has lower word counts and a narrower distribution.}}

\label{fig:static}
\end{figure*}

\rev{The evaluation reported in this manuscript comprises 6,422 image--text pairs and 10,703 unique referenced images, covering 15 clinical tasks and 27 named subtasks. As shown in Figure~\ref{fig:examples}, the records are organized into three formats: 1,100 Visual Question Answering (VQA) pairs that map an image--text query to a text response; 3,872 Image Editing pairs that require instruction-guided image modifications; and 1,450 Contextual Multimodal Generation pairs that require image and/or textual outputs. For reporting, we use six canonical modality labels: CT, MRI, Ultrasound, X-ray, Pathology, and Clinical Photography. Records whose source metadata cannot be mapped reliably are retained in an ``other/unspecified'' category. Task statistics are visualized in Figure~\ref{fig:static}.}

\rev{The design of \bench uses image-grounded instructions and open-ended outputs. We use \textit{contextual entanglement} as an operational design term for the dependence of an instruction's intended answer or transformation on the particular image instance, rather than on the task wording alone. Section~\ref{sec:instruction_entanglement} evaluates this dependence through semantic-alignment and behavioral pairing-sensitivity analyses.}

\rev{To adapt checklist-guided evaluation~\cite{zhang2025llmevalmed} across tasks, we define $\mathcal{K}_t=\mathcal{K}_{\mathrm{shared}}\cup\mathcal{K}_{\mathrm{task}}(t)$. The shared component comprises five clinical dimensions: Anatomical Accuracy (AA), Clinical-Finding Accuracy (CFA), and Instruction Compliance (IC), which apply across formats; and Cross-Modal Consistency (CMC) and Hallucination/Omission Control (HOC), which apply only to contextual multimodal generation. The task component specifies format-specific requirements. Lingshu-32B~\cite{lingshu2025} verifies the requirements before assigning ratings and a separate raw holistic score.}

\rev{The same task-specific checklist is applied under two evidence settings. In the reference-aware setting, the judge is additionally provided with the reference image or answer reviewed by clinical experts and GPT-4o. In the reference-free setting, the judge uses only the input, instruction, and candidate output. The checklist functions as an internal, prompt-guided verification procedure rather than producing checklist-level metrics. Instead, the judge assigns a rating and provides brief supporting evidence for each applicable dimension. Failure to meet a core requirement has a dominant effect on the 1--5 rating. The scale assigns 5 to outputs that fully satisfy the requirements, 4 to substantively correct outputs with only minor issues, 3 to partially complete or uncertain outputs, 2 to outputs with a major failure on a core requirement, and 1 to irrelevant, critically incorrect, or unusable outputs.}

\rev{We report SSIM~\cite{wang2004ssim}, PSNR~\cite{hore2010}, and LPIPS~\cite{lpips} as reference-based image-fidelity measures, and BLEU~\cite{papineni2002bleu}, PubMedBERTScore~\cite{zhang2019bertscore,PubMedBERT}, and Exact Match (EM)~\cite{squad2016} as text-agreement measures. EM is not reported for Gen. or Edit. because these tasks do not have a single exact answer. Checklist-based evidence is retained for error analysis, while the main table separately reports the raw holistic scores produced under the reference-aware and reference-free settings.}

\rev{The evaluation spans six modalities and fifteen heterogeneous task types without a shared annotation ontology based on pixel-aligned masks, bounding boxes, points, or surfaces. It evaluates full-image annotation and pseudo-segmentation outputs through reference-based fidelity and representation metrics together with a structured clinical review rubric. Spatial localization evaluation is reserved for a separate extension supported by expert spatial annotations.}

\subsection{Dataset Collection and Preprocessing}
\label{Dataset Collection and Preprocessing}
\color{black}

\rev{This section describes the collection and preprocessing of source medical images, comprising two components: (1) dataset collection and (2) two-stage filtering.}

\rev{\textbf{Dataset Collection.}}
\rev{To support broad and diverse evaluation coverage, we curated medical images from a wide range of public datasets and online repositories. These images span multiple modalities, including CT, MRI, ultrasound, X-ray, histopathology, and clinical photography.}

\rev{\textbf{Two-stage Filtering.}}
\rev{The preprocessing pipeline uses a two-stage filtering strategy to select task-relevant medical images. First, a coarse-grained, rule-based filter is applied to the source metadata to identify candidate images that satisfy task-specific metadata criteria. GPT-4o~\cite{achiam2023gpt} then performs semantic filtering based on both the candidate images and their associated metadata. Only images that meet the requirements of the target task are retained.}

\subsection{Image Pair Synthesis}
\label{Image Pair Synthesis}
\color{black}

To synthesize task-specific image pairs, we begin with a medical image corpus $\mathcal{D} = \{(\boldsymbol{X}_i, \boldsymbol{M}_i)\}_{i=1}^K$, where each image $\boldsymbol{X}_i$ is accompanied by metadata $\boldsymbol{M}_i$ containing domain-specific annotations such as modality, anatomical labels, and pathology tags. Our synthesis pipeline first uses $\boldsymbol{M}_i$ to filter $\mathcal{D}$ and select source images $\boldsymbol{X}_{\text{source}}$ that are suitable for specific tasks. To generate target images $\boldsymbol{X}_{\text{target}}$, we then apply a set of image transformations $\mathcal{T}$ to $\boldsymbol{X}_{\text{source}}$.

These transformations are categorized based on \mbox{task requirements}:

\setcounter{dbltopnumber}{1}
\AddToHookNext{shipout/after}{\AddToHookNext{shipout/after}{\AddToHookNext{shipout/after}{\setcounter{dbltopnumber}{4}}}}
\begin{table*}[!t]
\centering
\caption{\rev{Three-tier evaluation framework for separating auxiliary reference-based similarity from clinically oriented assessment. \textbf{w. GT} and \textbf{w.o. GT} denote reference-aware and reference-free views. AA, CFA, IC, CMC, and HOC denote anatomical accuracy, clinical-finding accuracy, instruction compliance, cross-modal consistency, and hallucination/omission control.}}
\label{tab:eval_metrics_summary}
\footnotesize
\setlength{\tabcolsep}{3pt}
\renewcommand{\arraystretch}{1.06}
\resizebox{1\linewidth}{!}{ 
\begin{tabular}{@{}m{0.105\textwidth}m{0.17\textwidth}m{0.655\textwidth}@{}}
\toprule[1.5pt]
\rev{\textbf{Eval Level}} & \rev{\textbf{Metric}} & \rev{\textbf{Description}} \\ 
\midrule
\multirow{4}{*}{\rev{\textbf{Image}}} 
 & \rev{SSIM~\cite{MultiscaleSSIM}} & \rev{Full-image structural fidelity to the reference.} \\
 & \rev{PSNR~\cite{hore2010}} & \rev{Pixel-wise reconstruction fidelity to the reference.} \\
 & \rev{LPIPS~\cite{lpips}} & \rev{Deep-feature perceptual distance (lower is better). These three measures describe global visual fidelity only and are not clinical-correctness measures.} \\
 & \rev{\makecell[l]{MedImageInsight\\Similarity~\cite{medimageinsight}}} & \rev{Cosine similarity between MedImageInsight embeddings of the generated and reference images.} \\
\midrule
\multirow{3}{*}{\rev{\textbf{Text}}} 
 & \rev{BLEU~\cite{papineni2002bleu}} & \rev{Reference-based lexical similarity.} \\
 & \rev{BERTScore~\cite{zhang2019bertscore}} & \rev{Reference-based PubMedBERT semantic similarity.} \\
 & \rev{EM~\cite{squad2016}} & \rev{Normalized exact agreement between the generated and reference answers.} \\
\midrule
\multirow{9}{*}{\rev{\makecell[c]{\textbf{Clinical}\\\textbf{holistic}}}} 
 & \multicolumn{2}{@{}m{0.845\textwidth}@{}}{\rev{Checklist-guided Lingshu-32B judge~\cite{lingshu2025}, with each dimension scored on a 1--5 scale.}} \\
\cmidrule(l){2-3}
 & \multicolumn{2}{@{}m{0.845\textwidth}@{}}{\rev{\textbf{w. GT:} Compares the generated output with the image/text reference.}} \\
 & \multicolumn{2}{@{}m{0.845\textwidth}@{}}{\rev{\textbf{w.o. GT:} Assesses the generated output using only the input image/instruction.}} \\
\cmidrule(l){2-3}
 & \multicolumn{2}{@{}m{0.845\textwidth}@{}}{\rev{\textit{Five dimensions (CMC and HOC apply only to multimodal generation because image-editing outputs do not include paired text.)}}} \\
 & \rev{AA} & \rev{Anatomy, site, and laterality are plausible and match the evidence.} \\
 & \rev{CFA} & \rev{Findings, disease status, and fine clinical details agree with the evidence.} \\
 & \rev{IC} & \rev{The requested transformation or answer is completed as intended.} \\
 & \rev{CMC} & \rev{Generated image and text support the same clinical content.} \\
 & \rev{HOC} & \rev{Penalizes invented or clinically relevant missing content.} \\
\bottomrule[1.5pt]
\end{tabular}%
}
\end{table*}

\begin{enumerate}
    \item \rev{Rule-based Transforms:} Deterministic operations using classical image processing tools. 
    \item \rev{Model-based Transforms:} Semantic and stylistic edits performed by advanced generative models, guided by textual instructions or image features.
\end{enumerate}

\rev{The organ-removal task is included as a controlled counterfactual inpainting task designed to test whether an image-editing model can remove an anatomical structure while preserving the surrounding anatomy. This paradigm is used to probe shortcut reliance in medical image interpretation~\cite{cooke2026roentmod} and is not intended as a clinical preprocessing operation.}

\rev{This process yields a pool of candidate image pairs $(\boldsymbol{X}_{\text{in}}, \boldsymbol{X}_{\text{out}})$. Depending on the task definition, $\boldsymbol{X}_{\text{source}}$ and $\boldsymbol{X}_{\text{target}}$ are assigned to $\boldsymbol{X}_{\text{in}}$ and $\boldsymbol{X}_{\text{out}}$, respectively, or with the assignment reversed. Released pairs undergo clinical-expert and GPT-4o review for anatomical accuracy, pathological plausibility, and clinical coherence.}

\subsection{Text Pair Synthesis}
\label{Text Pair Synthesis}
\color{black}

To pair each candidate image transformation with aligned textual supervision, we produce a corresponding instruction $\boldsymbol{T}$ and answer $\boldsymbol{A}$ for every candidate image pair $(\boldsymbol{X}_{\text{in}}, \boldsymbol{X}_{\text{out}})$.

\textbf{Semantic Extraction.}
To convert each candidate image pair \((\boldsymbol{X}_{\text{in}}, \boldsymbol{X}_{\text{out}})\) and its metadata into a task-specific textual representation, we use Qwen3-VL to extract structured semantic information $\boldsymbol{\mathcal{M}} = \{m_1, m_2, \dots, m_L\}$. To capture the information needed for downstream template selection, the extraction follows four key principles: (a) Basic Information Provision, (b) Detailed Comparison, (c) Structure-Level Discrepancy, and (d) Core Content Summary. We identify task-specific templates from a predefined template library $\mathcal{T}_{\text{task}}$ using the semantic representation $\boldsymbol{\mathcal{M}}$ and fill their placeholders with elements from $\boldsymbol{\mathcal{M}}$. This produces the raw instruction--answer pair $\boldsymbol{\mathcal{R}}_{\text{raw}}=(\boldsymbol{T}_{\text{raw}},\boldsymbol{A}_{\text{raw}})$, formalized as follows:

\begin{equation}
\boldsymbol{\mathcal{R}}_{\text{raw}}
= (\boldsymbol{T}_{\text{raw}},\boldsymbol{A}_{\text{raw}})
= \boldsymbol{\phi}(\boldsymbol{\mathcal{M}},\mathcal{T}_{\text{task}}).
\end{equation}

\textbf{Contextual Augmentation.}
To strengthen \textit{contextual entanglement} and precisely align visual content with textual descriptions, we refine $\boldsymbol{\mathcal{R}}_{\text{raw}}$ through a refinement function $\boldsymbol{\psi}$ that leverages GPT-4o {\color{black}\cite{achiam2023gpt}}. This function integrates the input--output image pair $(\boldsymbol{X}_{\text{in}}, \boldsymbol{X}_{\text{out}})$, the extracted metadata $\boldsymbol{\mathcal{M}}$, and the raw instruction--answer pair $\boldsymbol{\mathcal{R}}_{\text{raw}}$ to generate a clinically accurate and linguistically diverse final instruction $\boldsymbol{T}$ along with its \mbox{corresponding answer $\boldsymbol{A}$}:

\begin{equation}
(\boldsymbol{T}, \boldsymbol{A}) = \boldsymbol{\psi}\big(\boldsymbol{X}_{\text{in}}, \boldsymbol{X}_{\text{out}}, \boldsymbol{\mathcal{M}}, \boldsymbol{\mathcal{R}}_{\text{raw}}\big).
\end{equation}

To enhance semantic fidelity, promote linguistic diversity, and improve robustness across varied styles of clinical expression, the refinement process applies synonym substitution, syntactic rephrasing, and domain-specific terminology injection.

\newpage
\subsection{Postprocessing}
\label{Postprocessing}
\color{black}

\textbf{Automatic Quality Inspection.}
To facilitate VLM-based automatic review, we conduct \textit{Image Grounding} preprocessing by adding unobtrusive text identifiers to $\boldsymbol{X}_{\text{in}}$ and $\boldsymbol{X}_{\text{out}}$. We then submit the annotated image pair, instruction \(\boldsymbol{T}\), and reference answer \(\boldsymbol{A}\) to GPT-4o. To assess the consistency and quality of the model output relative to the ground truth, a specialized prompt guides the evaluation.

\setcounter{dbltopnumber}{1}
\begin{table*}[htb]
\centering
\vspace*{-0.75em}
\caption{\rev{Main results at \bench. To compare systems across complementary evidence views, we report reference-based fidelity and similarity measures together with structured holistic assessment. \textbf{Gen.}, \textbf{Edit.}, and \textbf{VQA} denote multimodal generation, image editing, and visual question answering. Holistic scores (1--5) are from the checklist-guided Lingshu-32B~\cite{lingshu2025} judge, while \textbf{w. GT} and \textbf{w.o. GT} denote reference-aware and reference-free views. MII, PBScore, and EM denote MedImageInsight similarity, PubMedBERTScore, and Exact Match, respectively. Metrics in $[0,1]$ are reported as percentages, and lower LPIPS is better. EM applies only to VQA. Bold and underline mark the best and second-best results.}}
\label{tab:Main_Result}
\revisionon
\scriptsize
\renewcommand{\arraystretch}{1.00}
\resizebox{1.0\textwidth}{!}{%
 \tabcolsep=0.1cm
\begin{tabular}{@{}llccccccccc@{}}
\toprule[1.5pt]
\multicolumn{2}{c|}{\multirow{2}{*}{\textbf{System}}} &
\multicolumn{2}{c|}{\textbf{Holistic (1--5)}} &
\multicolumn{4}{c|}{\textbf{Image}} &
\multicolumn{3}{c}{\textbf{Text}} \\
\multicolumn{2}{c|}{} &
\textbf{w. GT $\uparrow$} & \multicolumn{1}{c|}{\textbf{w.o. GT $\uparrow$}} &
\textbf{SSIM $\uparrow$} & \textbf{PSNR $\uparrow$} & \textbf{LPIPS $\downarrow$} &
\multicolumn{1}{c|}{\textbf{MII $\uparrow$}} &
\textbf{BLEU $\uparrow$} & \textbf{PBScore $\uparrow$} & \textbf{EM $\uparrow$} \\
\midrule
\multicolumn{1}{l|}{\multirow{8}{*}{\rotatebox{90}{\textbf{Gen.}}}} & \multicolumn{1}{l|}{\unifiedmodel Gemini-2.5-flash-image~\cite{comanici2025gemini25pushingfrontier}} & \textbf{2.90} & \multicolumn{1}{c|}{\textbf{3.48}} & \textbf{37.61} & \textbf{11.85} & \textbf{46.15} & \multicolumn{1}{c|}{\textbf{65.36}} & \underline{0.61} & \textbf{88.95} & \textemdash \\
\multicolumn{1}{l|}{} & \multicolumn{1}{l|}{\unifiedmodel Show-o~\cite{xie2024showo}} & 1.04 & \multicolumn{1}{c|}{1.04} & 26.94 & \underline{9.72} & 69.99 & \multicolumn{1}{c|}{33.92} & 0.53 & 86.07 & \textemdash \\
\multicolumn{1}{l|}{} & \multicolumn{1}{l|}{\unifiedmodel Ming-UniVision~\cite{huang2025mingunivisionjointimageunderstanding}} & 2.02 & \multicolumn{1}{c|}{\underline{2.52}} & \underline{33.82} & 8.93 & \underline{59.70} & \multicolumn{1}{c|}{\underline{54.64}} & 0.57 & 87.54 & \textemdash \\
\multicolumn{1}{l|}{} & \multicolumn{1}{l|}{\compositional Qwen3-VL \& Seedream-4.0~\cite{yang2025qwen3,Seedream}} & \underline{2.18} & \multicolumn{1}{c|}{2.23} & 17.61 & 8.83 & 64.28 & \multicolumn{1}{c|}{47.47} & 0.58 & 88.11 & \textemdash \\
\multicolumn{1}{l|}{} & \multicolumn{1}{l|}{\compositional Qwen3-VL \& DALL-E~3~\cite{yang2025qwen3,betker2023improving}} & 1.59 & \multicolumn{1}{c|}{1.63} & 13.34 & 9.20 & 66.12 & \multicolumn{1}{c|}{44.45} & 0.59 & 88.12 & \textemdash \\
\multicolumn{1}{l|}{} & \multicolumn{1}{l|}{\compositional Qwen3-VL \& Imagen-4.0~\cite{yang2025qwen3,google2025imagen4}} & 2.09 & \multicolumn{1}{c|}{2.24} & 22.12 & 9.56 & 62.84 & \multicolumn{1}{c|}{46.94} & 0.59 & 88.08 & \textemdash \\
\multicolumn{1}{l|}{} & \multicolumn{1}{l|}{\compositional GPT-4o \& Seedream-4.0~\cite{achiam2023gpt,Seedream}} & 1.87 & \multicolumn{1}{c|}{1.86} & 16.67 & 9.01 & 64.80 & \multicolumn{1}{c|}{46.62} & \textbf{0.64} & 87.80 & \textemdash \\
\multicolumn{1}{l|}{} & \multicolumn{1}{l|}{\compositional Gemini-2.5-pro \& Seedream-4.0~\cite{comanici2025gemini25pushingfrontier,Seedream}} & 2.07 & \multicolumn{1}{c|}{2.21} & 17.10 & 8.64 & 64.68 & \multicolumn{1}{c|}{47.64} & \underline{0.61} & \underline{88.23} & \textemdash \\
\midrule
\multicolumn{1}{l|}{\multirow{10}{*}{\rotatebox{90}{\textbf{Edit.}}}} & \multicolumn{1}{l|}{\unifiedmodel Gemini-2.5-flash-image~\cite{comanici2025gemini25pushingfrontier}} & \underline{3.85} & \multicolumn{1}{c|}{\underline{4.19}} & \underline{47.24} & \underline{15.74} & \underline{26.67} & \multicolumn{1}{c|}{\underline{82.55}} & \textemdash & \textemdash & \textemdash \\
\multicolumn{1}{l|}{} & \multicolumn{1}{l|}{\unifiedmodel Show-o~\cite{xie2024showo}} & 1.07 & \multicolumn{1}{c|}{1.08} & 26.33 & 9.65 & 70.15 & \multicolumn{1}{c|}{33.71} & \textemdash & \textemdash & \textemdash \\
\multicolumn{1}{l|}{} & \multicolumn{1}{l|}{\unifiedmodel Ming-UniVision~\cite{huang2025mingunivisionjointimageunderstanding}} & 3.08 & \multicolumn{1}{c|}{3.36} & 46.06 & 14.97 & 34.71 & \multicolumn{1}{c|}{77.03} & \textemdash & \textemdash & \textemdash \\
\multicolumn{1}{l|}{} & \multicolumn{1}{l|}{\unifiedmodel Seedream-4.0~\cite{Seedream}} & 1.62 & \multicolumn{1}{c|}{1.50} & 18.16 & 7.98 & 64.29 & \multicolumn{1}{c|}{46.37} & \textemdash & \textemdash & \textemdash \\
\multicolumn{1}{l|}{} & \multicolumn{1}{l|}{\unifiedmodel Qwen-Image-Edit~\cite{wu2025qwenimage}} & 3.77 & \multicolumn{1}{c|}{4.11} & \textbf{53.10} & \textbf{16.46} & \textbf{24.57} & \multicolumn{1}{c|}{\textbf{83.88}} & \textemdash & \textemdash & \textemdash \\
\multicolumn{1}{l|}{} & \multicolumn{1}{l|}{\compositional Qwen3-VL \& GPT-image-1~\cite{yang2025qwen3,openai2025gptimage}} & \textbf{4.02} & \multicolumn{1}{c|}{\textbf{4.37}} & 35.20 & 12.06 & 47.29 & \multicolumn{1}{c|}{72.54} & \textemdash & \textemdash & \textemdash \\
\multicolumn{1}{l|}{} & \multicolumn{1}{l|}{\compositional Qwen3-VL \& DALL-E~3~\cite{yang2025qwen3,betker2023improving}} & 1.59 & \multicolumn{1}{c|}{1.50} & 15.17 & 8.32 & 64.39 & \multicolumn{1}{c|}{51.73} & \textemdash & \textemdash & \textemdash \\
\multicolumn{1}{l|}{} & \multicolumn{1}{l|}{\compositional Qwen3-VL \& Imagen-4.0~\cite{yang2025qwen3,google2025imagen4}} & 2.87 & \multicolumn{1}{c|}{2.94} & 24.78 & 8.98 & 58.06 & \multicolumn{1}{c|}{59.51} & \textemdash & \textemdash & \textemdash \\
\multicolumn{1}{l|}{} & \multicolumn{1}{l|}{\compositional GPT-4o \& Imagen-4.0~\cite{achiam2023gpt,google2025imagen4}} & 2.60 & \multicolumn{1}{c|}{2.64} & 24.14 & 8.82 & 59.31 & \multicolumn{1}{c|}{56.88} & \textemdash & \textemdash & \textemdash \\
\multicolumn{1}{l|}{} & \multicolumn{1}{l|}{\compositional Gemini-2.5-pro \& Imagen-4.0~\cite{comanici2025gemini25pushingfrontier,google2025imagen4}} & 2.65 & \multicolumn{1}{c|}{2.80} & 24.86 & 8.84 & 58.64 & \multicolumn{1}{c|}{58.78} & \textemdash & \textemdash & \textemdash \\
\midrule
\multicolumn{1}{l|}{\multirow{7}{*}{\rotatebox{90}{\textbf{VQA}}}} & \multicolumn{1}{l|}{\unifiedmodel Qwen3-VL-235B-A22B-Instruct~\cite{yang2025qwen3}} & \textbf{4.04} & \multicolumn{1}{c|}{\textbf{4.55}} & \textemdash & \textemdash & \textemdash & \multicolumn{1}{c|}{\textemdash} & 2.21 & 87.67 & \textbf{43.28} \\
\multicolumn{1}{l|}{} & \multicolumn{1}{l|}{\unifiedmodel Gemini-2.5-pro~\cite{comanici2025gemini25pushingfrontier}} & \underline{4.00} & \multicolumn{1}{c|}{\underline{4.50}} & \textemdash & \textemdash & \textemdash & \multicolumn{1}{c|}{\textemdash} & 2.22 & \underline{87.77} & 18.10 \\
\multicolumn{1}{l|}{} & \multicolumn{1}{l|}{\unifiedmodel GPT-4o~\cite{achiam2023gpt}} & 3.70 & \multicolumn{1}{c|}{4.14} & \textemdash & \textemdash & \textemdash & \multicolumn{1}{c|}{\textemdash} & \underline{2.25} & 87.36 & 24.67 \\
\multicolumn{1}{l|}{} & \multicolumn{1}{l|}{\unifiedmodel HuatuoGPT-Vision~\cite{PubMedVision}} & 3.61 & \multicolumn{1}{c|}{4.03} & \textemdash & \textemdash & \textemdash & \multicolumn{1}{c|}{\textemdash} & \textbf{3.79} & \textbf{87.86} & \underline{28.37} \\
\multicolumn{1}{l|}{} & \multicolumn{1}{l|}{\unifiedmodel RadFM~\cite{radfm}} & 1.89 & \multicolumn{1}{c|}{1.95} & \textemdash & \textemdash & \textemdash & \multicolumn{1}{c|}{\textemdash} & 1.13 & 86.12 & 13.33 \\
\multicolumn{1}{l|}{} & \multicolumn{1}{l|}{\unifiedmodel Show-o~\cite{xie2024showo}} & 1.59 & \multicolumn{1}{c|}{1.64} & \textemdash & \textemdash & \textemdash & \multicolumn{1}{c|}{\textemdash} & 0.05 & 74.91 & 3.01 \\
\multicolumn{1}{l|}{} & \multicolumn{1}{l|}{\unifiedmodel Ming-UniVision~\cite{huang2025mingunivisionjointimageunderstanding}} & 2.82 & \multicolumn{1}{c|}{3.17} & \textemdash & \textemdash & \textemdash & \multicolumn{1}{c|}{\textemdash} & 1.62 & 87.08 & 25.68 \\
\bottomrule[1.5pt]
\end{tabular}%
}
\revisionoff
\end{table*}

\section{Experiments and Analysis}

\subsection{Evaluating MedGEN-Bench}

\rev{\textbf{Experiment Setup.}} 
\rev{We evaluate ten compositional frameworks that jointly produce textual and visual outputs. We also assess two dedicated image-editing models, Qwen-Image-Edit~\cite{wu2025qwenimage} and Seedream-4.0~\cite{Seedream}, as well as three unified understanding-and-generation models, Show-o~\cite{xie2024showo,xie2025showo2}, Gemini-2.5-flash-image~\cite{comanici2025gemini25pushingfrontier}, and Ming-UniVision~\cite{huang2025mingunivisionjointimageunderstanding}. The VQA evaluation includes three general-purpose vision-language models, Qwen3-VL-235B-A22B-Instruct~\cite{yang2025qwen3}, Gemini-2.5-pro~\cite{comanici2025gemini25pushingfrontier}, and GPT-4o~\cite{achiam2023gpt}, together with two medical vision-language models, HuatuoGPT-Vision~\cite{PubMedVision} and RadFM~\cite{radfm}. Table~\ref{tab:Main_Result} summarizes the main evaluation results.}




\rev{\textbf{Reference-based Results for the Image-output Formats.}}
\rev{Reference-based metrics indicate room for improvement in matching instance-specific target images across both image-output formats in \bench. Among the multimodal generation systems, Gemini-2.5-flash-image achieves the strongest reference-based image results, with 37.61\% SSIM, 11.85~dB PSNR, 46.15\% LPIPS, and 65.36\% MedImageInsight similarity. Among the image-editing systems, Qwen-Image-Edit achieves 53.10\% SSIM, 16.46~dB PSNR, 24.57\% LPIPS, and 83.88\% MedImageInsight similarity. The best editing SSIM is 0.5310 on its original scale, leaving room for improving structural fidelity and preserving instance-specific anatomy. Because the two formats differ in task mix, cases, evaluated systems, and reference images, these format-level aggregates are not used to rank their relative difficulty.}

\rev{\textbf{Within-block Configuration Differences Are Descriptive.}}
\rev{Table~\ref{tab:Main_Result} reports point estimates from separate generation and editing blocks. Seedream-4.0 generation rows vary across upstream VLMs, whereas Qwen3-VL editing rows vary across image backends. Different task mixes, cases, output formats, and configurations make these ranges descriptive rather than matched effects. They do not establish an independent or causal effect of either component, identify a primary bottleneck, or support a cross-block comparison. Such claims require factorial experiments with matched cases, repeated runs, and uncertainty estimates.}

\rev{\textbf{Medical Specialization Does Not Guarantee Strong Performance Across Heterogeneous VQA Tasks.}}
\rev{RadFM's weaker overall performance cannot be attributed simply to low semantic similarity with the reference answers. Its PBScore is 86.12\%, close to the range of 87.36\% to 87.77\% achieved by the three general-purpose VLMs. However, its EM is only 13.33\%, compared with the general-purpose VLM range of 18.10\% to 43.28\%. Its reference-aware and reference-free holistic scores are 1.89 and 1.95, respectively, also below the corresponding ranges of 3.70 to 4.04 and 4.14 to 4.55. This performance profile is therefore consistent with weaker exact-answer compliance and image-specific task grounding in practice despite relatively high semantic similarity to the reference answers. Without controlled comparisons of architecture, training data, prompting, and decoding, we do not attribute this gap to a single cause, including medical domain pretraining.}

\rev{\textbf{Reference Availability Affects Holistic Evaluation.}}
\rev{Although the same checklist is applied in both settings, reference availability changes the evidence directly provided to the judge. Averaged across the system-level means reported in Table~\ref{tab:Main_Result}, the reference-free scores exceed the reference-aware scores by 0.18, 0.14, and 0.33 points for generation, editing, and VQA, respectively. The reference-aware setting can consequently assign lower scores to candidates that disagree with the available answer or patient-specific target reviewed by clinical experts and GPT-4o, whereas the reference-free setting evaluates plausibility and instruction fulfillment using only the input, instruction, and candidate output, without an external target for comparison. The larger mean gap for VQA is consistent with plausible but reference-discordant answers receiving higher scores when the reference answer is unavailable. Occasional reversals among editing systems further show that the two independently elicited scores capture complementary evidence rather than follow a fixed order.}


\subsection{\rev{Contextual Entanglement: Semantic Alignment and Pairing Sensitivity}}
\label{sec:instruction_entanglement}

\rev{\textbf{Semantic Alignment of Contextual Instructions.}}
\rev{We first evaluate the semantic alignment of contextual instructions at the data-construction level using image--text embedding similarity. We randomly sample 1,000 image--instruction pairs and compare the image--text similarities obtained with the original templates and their contextually augmented instructions. For each pair, similarity gain is defined as the difference between the augmented instruction's similarity to the image and the average similarity of the comparison benchmark queries to the same image. Pass rate is defined as the proportion of pairs for which the augmented instruction exceeds this benchmark-query average. By comparing the original and augmented instructions against the same image, this analysis isolates the contribution of contextual detail to semantic alignment. The paired comparison keeps the image fixed, so changes in similarity can be attributed to the instruction wording.}

\begin{table*}[!t]
\centering
\caption{\rev{Behavioral pairing sensitivity on a random 1,000-case sample spanning all benchmark tasks and canonical modalities. The table reports reference-free decreases, $\Delta=S_{\mathrm{correct}}-S_{\mathrm{control}}$, on five dimensions scored by Lingshu-32B~\cite{lingshu2025}. Larger values indicate stronger pairing dependence. The primary cross-modal-consistency endpoint includes a bootstrap 95\% confidence interval and Holm-adjusted paired Wilcoxon \textit{p} value.}}
\label{tab:entanglement_control}
\revisionon
{\fontsize{8}{8.5}\selectfont
\setlength{\tabcolsep}{2.5pt}
\renewcommand{\arraystretch}{1.00}
\begin{tabular*}{\linewidth}{@{\extracolsep{\fill}}lccccccc@{}}
\toprule[1.5pt]
\multirow{2}{*}{\textbf{Control}} & \multicolumn{2}{c}{\textbf{Cross-modal}} & \multirow{2}{*}{\textbf{Anat.}} & \multirow{2}{*}{\textbf{Findings}} & \multirow{2}{*}{\textbf{Compliance}} & \multirow{2}{*}{\makecell{\textbf{Halluc. /}\\\textbf{omission}}} & \multirow{2}{*}{\textbf{$p_{\mathrm{Holm}}$}} \\
\cmidrule(lr){2-3}
& \textbf{$\Delta$} & \textbf{95\% CI} & & & & & \\
\midrule
Blank image & 2.16 & 1.88--2.44 & 1.34 & 2.10 & 2.10 & 2.14 & $2.23\!\times\!10^{-24}$ \\
Unrelated image & 1.07 & 0.81--1.33 & 0.32 & 0.78 & 0.43 & 0.39 & $1.43\!\times\!10^{-11}$ \\
Same-modality image & 0.57 & 0.37--0.80 & 0.16 & 0.42 & 0.20 & 0.16 & $1.49\!\times\!10^{-6}$ \\
Image shuffle & 0.64 & 0.43--0.86 & 0.28 & 0.65 & 0.28 & 0.30 & $4.60\!\times\!10^{-7}$ \\
\textbf{Instruction shuffle} & \textbf{0.52} & \textbf{0.31--0.72} & \textbf{0.12} & \textbf{0.49} & \textbf{0.25} & \textbf{0.23} & \textbf{$3.44\!\times\!10^{-6}$} \\
\bottomrule[1.5pt]
\end{tabular*}
}
\par\vspace{0pt}\noindent
\begin{minipage}{\linewidth}
\vspace{0.6em}
\fontsize{9}{12}\selectfont\rev{\textit{Note:} Scores use the 1--5 rubric. The anatomical, findings, and compliance columns summarize all formats. The cross-modal and hallucination/omission columns use contextual multimodal-generation cases only. $p_{\mathrm{Holm}}$ refers to the cross-modal-consistency endpoint, and incomplete judge scores are excluded pairwise. The instruction-shuffle row is bolded for readability, while its non-cross-modal values are auxiliary exploratory diagnostics.}
\end{minipage}

\revisionoff
\end{table*}

\definecolor{maincaseblue}{RGB}{0,98,155}
\definecolor{maincasepositive}{RGB}{46,125,82}
\definecolor{maincasenegative}{RGB}{174,55,55}
\ifnum\MarkedUpCopy=1
  \colorlet{maincasebody}{\revisioncolor}
\else
  \colorlet{maincasebody}{black}
\fi

\newtcolorbox{maincasecard}[1]{
  enhanced,
  colback=white,
  colframe=maincaseblue,
  colbacktitle=maincaseblue,
  coltitle=white,
  coltext=maincasebody,
  fonttitle=\sffamily\bfseries\small,
  title={#1},
  boxrule=0.65pt,
  arc=1.2mm,
  left=5pt,
  right=5pt,
  top=4pt,
  bottom=4pt,
  boxsep=0pt,
  before skip=5pt,
  after skip=5pt
}

\newtcolorbox{mainformatcontrastpanel}[2]{
  enhanced,
  colback=#2!4,
  colframe=#2,
  colbacktitle=#2,
  coltitle=white,
  coltext=maincasebody,
  fonttitle=\sffamily\bfseries\footnotesize,
  title={#1},
  boxrule=0.55pt,
  arc=0.8mm,
  left=4pt,
  right=4pt,
  top=3pt,
  bottom=3pt,
  boxsep=0pt
}

\newcommand{\maincasefield}[2]{\par\smallskip\noindent\textbf{#1.} #2}
\newcommand{\maingoodtext}[1]{\textcolor{maincasepositive!85!black}{\bfseries #1}}
\newcommand{\mainbadtext}[1]{\textcolor{maincasenegative!90!black}{\bfseries #1}}

\rev{\textbf{Improved Image--Instruction Alignment after Augmentation.}}
\rev{Figure~\ref{fig:entanglement} shows a rightward shift in image--instruction similarity following contextual augmentation. The mean similarity increases from 0.273 to 0.372, while the pass rates across the six benchmark comparisons range from 70.86\% to 96.45\%. These positive gains indicate that the augmented instructions exhibit stronger semantic alignment with their paired images than the comparison benchmark queries. Across the comparison benchmarks, the improvement is not concentrated in a single source of reference instructions. The higher pass rates show that contextual augmentation more consistently preserves the visual information needed for the paired task. Together, these results provide a clear basis for testing sensitivity to changes in the image and instruction pairing. This construction-level evidence motivates the behavioral pairing-sensitivity analysis that follows.}

\begin{minipage}{\textwidth}
\centering
\captionof{table}{\rev{Agreement between Lingshu-32B~\cite{lingshu2025} and medical-expert ratings. OHS is the overall score, and five shared dimensions include AA, CFA, and IC, which apply across formats, and CMC and HOC, which apply only to contextual multimodal generation.}}
\label{tab:researcher_audit_agreement}
\footnotesize
\setlength{\tabcolsep}{0pt}
\renewcommand{\arraystretch}{1.15}
\begin{tabular}{@{}l@{\hspace{2.5em}}c@{\hspace{2.5em}}c@{\hspace{2.5em}}c@{\hspace{2.5em}}c@{}}
\toprule[1.5pt]
\textbf{Dimension} & \textbf{MAE $\downarrow$} & \textbf{Pearson $r\uparrow$} & \textbf{Spearman $\rho\uparrow$} & \textbf{QWK $\uparrow$} \\
\midrule
OHS & 0.95 & 0.632 & 0.611 & 0.595 \\
AA & 1.02 & 0.565 & 0.531 & 0.516 \\
CFA & 1.19 & 0.553 & 0.505 & 0.484 \\
IC & 1.04 & 0.564 & 0.571 & 0.543 \\
CMC & 0.99 & 0.658 & 0.658 & 0.586 \\
HOC & 1.21 & 0.550 & 0.541 & 0.478 \\
\bottomrule[1.5pt]
\end{tabular}
\end{minipage}
\par\vspace{0.4em}

\noindent
\begin{minipage}[t]{0.465\textwidth}
\vspace{0pt}
\centering
\refstepcounter{figure}
\label{fig:entanglement}
\label{fig:ablation}
\includegraphics[width=\linewidth]{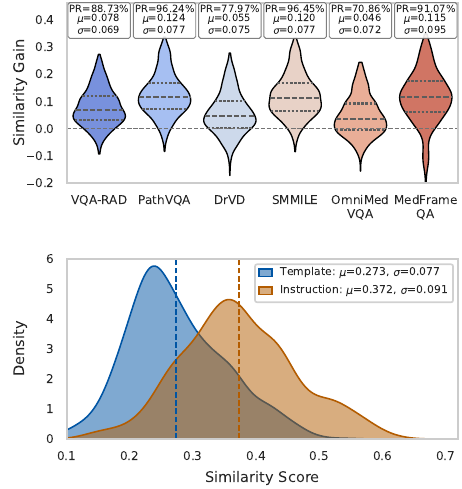}
\vspace*{-0.25cm}
\captionof*{figure}{\textbf{Figure~\thefigure}\enspace\rev{Semantic-alignment analysis. The top panel reports gains against six VQA benchmarks, and the bottom panel compares templates with augmented instructions. PR denotes pass rate, while $\mu$ and $\sigma$ denote mean and standard deviation.}}
\end{minipage}\hfill
\begin{minipage}[t]{0.51\textwidth}
\vspace{0pt}
\rev{\textbf{Behavioral Sensitivity to Image--Instruction Pairing.}}
\label{sec:entanglement_control}
\rev{Table~\ref{tab:entanglement_control} summarizes behavioral pairing-sensitivity results. To complement the semantic-alignment analysis in Figure~\ref{fig:entanglement}, we use the same random sample of 1,000 image--instruction pairs spanning VQA, image editing, and contextual multimodal generation. We evaluate them under six conditions: correct pairing, blank image, unrelated image, same-modality replacement, image shuffle, and instruction shuffle. Lingshu-32B~\cite{lingshu2025} scores anatomical accuracy, clinical-finding accuracy, and instruction compliance for all formats, plus cross-modal consistency and hallucination/omission control for contextual multimodal-generation outputs under reference-aware and reference-free settings. For each control condition, we define paired score decrease as $S_{\mathrm{correct}}-S_{\mathrm{control}}$ and report bootstrap 95\% confidence intervals and paired Wilcoxon signed-rank tests with Holm correction for multiple comparisons.}

\rev{Judge completeness ranges from 98\% to 100\%, and incomplete scores are excluded pairwise rather than scored zero.} The pre-specified primary endpoint is the reference-free decrease in cross-modal consistency under instruction shuffle within the contextual multimodal-generation subset. The observed decrease was 0.52, with a bootstrap 95\% CI of 0.31 to 0.72 and a Holm-adjusted paired Wilcoxon \textit{p} value of $3.44\times10^{-6}$.
\end{minipage}
\par\smallskip

For other controls, paired decreases were 2.16 for blank images, 1.07 for unrelated images, 0.64 for image shuffle, and 0.57 for same-modality replacement. Together, results link semantic-alignment gains in Figure~\ref{fig:entanglement} to behavioral sensitivity to perturbations of intended image--instruction pairing.
\par\smallskip

\rev{\textbf{Medical-expert Audit of the VLM Judge.}}
\rev{Table~\ref{tab:researcher_audit_agreement} compares Lingshu-32B~\cite{lingshu2025} with ratings from three clinicians on 1,000 cases stratified by task, format, and modality. Appendix~\ref{app:review} describes the sampling and reviewer specialties. Clinicians from radiology, ultrasonography, and pathology use a shared 1 to 5 rubric. Anatomical accuracy, clinical finding accuracy, and instruction compliance are assessed across formats, while cross modal consistency and hallucination or omission control are assessed for contextual multimodal generation. The judge produces the overall holistic score independently rather than calculating it as an arithmetic mean of the dimension scores. This design tests holistic agreement and its supporting clinical checks. Four complementary discrepancy, association, and ordinal-agreement statistics are reported. Pearson's $r$ measures linear correspondence. Spearman's $\rho$ measures preservation of case ranking. Quadratic weighted Cohen's $\kappa$ measures ordinal agreement while assigning more weight to larger score separations. For the overall holistic score, MAE is $0.95$, Pearson's $r$ is $0.632$, Spearman's $\rho$ is $0.611$, and quadratic weighted Cohen's $\kappa$ is $0.595$. The mean difference is below one point on the shared scale, and cases judged more favorably by clinicians generally also receive higher judge scores. The positive correlation and ordinal agreement values therefore indicate moderate agreement rather than exact interchangeability. The dimension results identify both useful agreement and residual disagreement. Cross modal consistency, assessed only for contextual multimodal generation, has the strongest Pearson and Spearman correlations, both $0.658$, with MAE $0.99$ and $\kappa$ $0.586$. For the remaining reported dimensions, MAE ranges from $1.02$ to $1.21$, Pearson's $r$ from $0.550$ to $0.565$, Spearman's $\rho$ from $0.505$ to $0.571$, and $\kappa$ from $0.478$ to $0.543$.}

\subsection{\rev{Closed-Ended Formats}}
\label{sec:qualitative_openended}
\vspace{-0.35em}

\noindent
\begin{minipage}[t]{0.445\textwidth}
\vspace{0pt}
\rev{Figure~\ref{fig:closed_open_contrast} contrasts closed-form selection with an open-ended report on the same CT case. Thus, a narrowed answer space can mask diagnostic substitution, motivating open-ended evaluation alongside multiple-choice accuracy. It can therefore reveal clinically plausible substitutions hidden by supplied options. This paired example is illustrative rather than a controlled estimate of the effect of removing options. The requested response format also changes. Nonetheless, the contrast shows why selection of a supplied diagnosis and generation of an image-grounded report should be assessed as complementary capabilities, the former tests discrimination within a fixed answer set, whereas the latter exposes unsupported substitutions that a restricted option list can conceal. It also distinguishes genuine visual reasoning from unsupported diagnostic matching. This distinction supports reporting both outcomes when evaluating image interpretation.}
\end{minipage}\hfill
\begin{minipage}[t]{0.53\textwidth}
\vspace{-2.0em}
\centering
\refstepcounter{figure}
\label{fig:closed_open_contrast}
\begin{maincasecard}{Closed-Form Selection versus Open-Ended Report}
\scriptsize
\noindent\textbf{Model:} HuatuoGPT-Vision \quad \textbf{CT} \quad \textbf{MedFrameQA}

\smallskip
\begin{minipage}[t]{0.28\linewidth}
  \vspace{0pt}\centering
  \begingroup
  \setlength{\fboxsep}{1pt}\setlength{\fboxrule}{0.8pt}%
  \fcolorbox{maincaseblue}{white}{\includegraphics[width=0.90\linewidth,height=2.05cm,keepaspectratio]{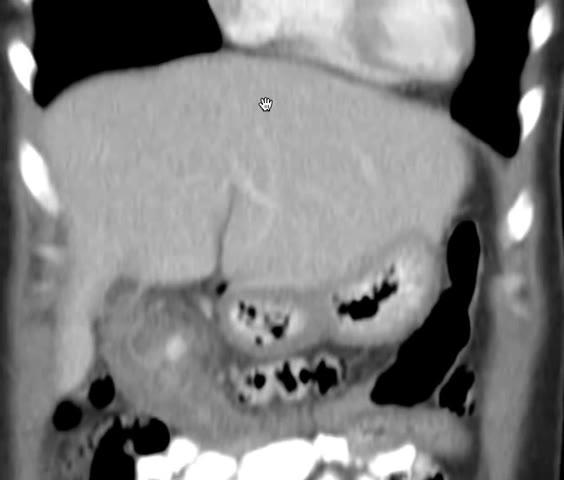}}%
  \endgroup\\[-0.2em]
  {\scriptsize\textcolor{maincaseblue}{\bfseries Same stored CT input}}
\end{minipage}\hfill
\begin{minipage}[t]{0.685\linewidth}
  \vspace{0pt}
  \maincasefield{Original question}{Which diagnosis best explains proximal transverse-colon diverticulitis together with disproportionate enlargement of the left hepatic lobe?}
  \maincasefield{Reference conclusion}{Cirrhosis with compensatory hypertrophy of the left hepatic lobe and proximal transverse-colon diverticulitis.}
\end{minipage}

\smallskip
\begin{mainformatcontrastpanel}{Closed-form VQA --- eight supplied options}{maincasepositive}
\textbf{Output:} \maingoodtext{``Option A: Cirrhosis with compensatory hypertrophy of the left lobe.'' This matches the supplied reference option.}
\end{mainformatcontrastpanel}

\smallskip
\begin{mainformatcontrastpanel}{Open-ended counterpart --- structured report}{maincasenegative}
\textbf{Output:} ``One possible diagnosis could be \mainbadtext{Budd--Chiari syndrome}.'' \mainbadtext{This replaces the reference cirrhosis conclusion with an unsupported alternative.}
\end{mainformatcontrastpanel}
\end{maincasecard}
\captionof*{figure}{\textbf{Figure~\thefigure}\enspace\rev{Open- versus closed-form contrast for stored HuatuoGPT-Vision outputs on the same CT slice and clinical question. The closed-form answer selects A, whereas the report response proposes Budd--Chiari syndrome instead of reference cirrhosis.}}
\end{minipage}
\par\vspace{-0.2em}

\section{Conclusion}
\rev{We introduce \bench, an open-ended benchmark of 6,422 reviewed image--text pairs across six modalities with 15 clinical tasks and 27 subtasks. It tests whether generated images and reports remain clinically and semantically aligned instead of evaluating them separately. Contextual augmentation increases mean image--instruction similarity from 0.273 to 0.372. Five pairing perturbations reduce cross-modal consistency, confirming sensitivity to intended image--instruction relations. A closed-form study shows that correct option selection may not yield an image-grounded report. Moderate VLM judge--expert agreement supports the rubric as complementary evidence rather than diagnostic validation. Future work will add longitudinal multi-view cases and federated multi-institutional evaluation of robustness and subgroup behavior without centralized data.

More broadly, these findings position \bench as a setting for asking whether an open-ended system is using clinical context, rather than merely producing a fluent description or selecting a plausible option. The paired construction makes it possible to compare the evidence carried by an image, the instruction that frames the task, and the report generated from their combination. It also exposes a critical difference between responses that remain aligned with the available evidence and responses that sound clinically credible while omitting a key finding or introducing an unsupported alternative. This distinction matters in medical generation, where surface coherence alone can conceal a failure of grounding. By making such mismatches measurable across formats and controls, the benchmark provides a common basis for comparing models, data curation strategies, prompting methods, and safety mechanisms under clinically relevant conditions.

Looking ahead, we view the benchmark as a foundation for evaluation that better reflects the longitudinal and collaborative nature of clinical care. Linked studies, richer patient histories, and reviews that account for uncertainty, evolving evidence, and institutional variation can extend \bench without losing its focus on image--instruction dependence. These additions can clarify when a model offers useful complementary evidence and when stronger expert oversight is necessary. They can also support more precise analyses of robustness, calibration, and subgroup behavior across sites. We hope \bench encourages development of medical generative systems that are not only fluent and technically capable, but also grounded, consistent, and responsive to the clinical context in which their outputs are used.}

\raggedbottom
\clearpage
\IEEEtriggeratref{90}
\bibliography{cites}

\clearpage
\par
\appendix
\raggedbottom
\newcommand{\appendixcontentsmajor}[2]{%
  \noindent\makebox[2.0em][l]{{\bfseries\ref*{#2}}}%
  \hyperref[#2]{{\color{metablue}\bfseries #1}}\hspace{0.25em}\dotfill\hspace{0.25em}%
  \makebox[2.0em][r]{\pageref*{#2}}\par\vspace{0.12em}%
}
\newcommand{\appendixcontentssub}[2]{%
  \noindent\hspace*{2.0em}\makebox[2.35em][l]{\ref*{#2}}%
  \hyperref[#2]{{\color{metablue}#1}}\hspace{0.25em}\dotfill\hspace{0.25em}%
  \makebox[2.0em][r]{\pageref*{#2}}\par\vspace{0.08em}%
}
\noindent{\large\bfseries Appendix\par}
\vspace{0.55em}
{\small
  \appendixcontentsmajor{Data Sources and Task Definitions}{app:data_tasks}
  \appendixcontentsmajor{Detailed Evaluation Protocol}{app:evaluation_protocol}
  \appendixcontentssub{Detailed Evaluation Metrics}{app:metrics}
  \appendixcontentssub{Image Fidelity and Medical-Image Representation}{app:image_fidelity}
  \appendixcontentssub{Text Similarity and Answer Agreement}{app:text_similarity}
  \appendixcontentssub{Clinical Holistic Assessment}{app:clinical_holistic}
  \appendixcontentssub{Aggregation and Error Analysis}{app:aggregation_error}
  \appendixcontentssub{Detailed Benchmark-Construction Expert Review}{app:review}
  \appendixcontentsmajor{Detailed Instruction-Entanglement Analysis}{app:ablation}
  \appendixcontentssub{Experimental Setup}{app:entanglement_setup}
  \appendixcontentssub{Similarity Computation}{app:similarity_computation}
  \appendixcontentssub{Evaluation Metrics}{app:entanglement_metrics}
  \appendixcontentssub{Semantic Alignment of Contextual Instructions}{app:semantic_alignment}
  \appendixcontentssub{Representative Examples}{app:representative_examples}
  \appendixcontentsmajor{Prompt Details and Examples}{app:prompt_details}
  \appendixcontentsmajor{Benchmark-Construction Prompt Examples}{app:construction_prompts}
}
\clearpage
\section{Data Sources and Task Definitions}
\label{app:data_tasks}

\begin{table}[H]
\centering
\vspace{-1.8em}
\caption{\rev{Source-data inventory provided to document the modality coverage of \bench. The table lists the contributing datasets under their canonical imaging modalities.}}
\label{tab:data_source}

\resizebox{0.75\textwidth}{!}{

\begin{tabular}{@{}ll@{}}
\toprule[1.5pt]
\textbf{Modality} & \textbf{Dataset Name} \\ \midrule
\multirow{8}{*}{\textbf{Radiography (X-ray)}} & ye2023sa\cite{ye2023sa}  \\
 & RSNA-Pneumonia\cite{RSNA-Pneumonia}  \\
 & MURA\cite{MURA}  \\
 & Covid19\_heywhale\cite{Covid19_heywhale}  \\
 & VinDr-SpineXR\cite{VinDr-SpineXR}  \\
 & Panoramic Dental Radiography / Tufts Dental Dataset\cite{Panoramic_Dental_Radiography} \\
 & SIIM-ACR\cite{SIIM-ACR} \\
 & MIMIC-CXR\cite{MIMIC-CXR} \\ \midrule
\multirow{13}{*}{\textbf{CT}} & CT-RATE\cite{CT-RATE} \\
 & DeepLesion\cite{DeepLesion} \\
 & COVID-CTset\cite{COVID-CTset} \\
 & LiTS (Liver Tumor Segmentation)\cite{LiTS} \\
 & zhou2024towards\cite{zhou2024towards} \\
 & StructSeg2019-subtask1\cite{StructSeg2019-subtask1} \\
 & KiTS 2021\cite{KiTS} \\
 & LNDb\cite{LNDb} \\
 & SARS-CoV-2 CT-scan\cite{SARS-CoV-2} \\
 & ye2023sa\cite{ye2023sa} \\
 & AMOS 2022\cite{AMOS} \\
 & PubMedVision\cite{PubMedVision} \\
 & RadImageNet\cite{RadImageNet} \\ \midrule
\multirow{6}{*}{\rev{\textbf{MRI}}} & BraTS 2020\cite{BraTS} \\
 & BraTS 2021\cite{BraTS} \\
 & LLD-MMRI\cite{LLD-MMRI} \\
 & MICCAI 2024 CARE MyoPS++\cite{Zhuang_2019,10159447,QIU2023102694,LI2023102808} \\
 & ISPY1-Tumor-SEG-Radiomics\cite{ISPY1-Tumor-SEG-Radiomics} \\
 & \rev{TotalSegmentator MRI\cite{TotalSegmentatorMRI}} \\ \midrule
\multirow{5}{*}{\textbf{Ultrasound}} & TG3K\cite{TG3K} \\
 & Annotated Ultrasound Liver Images\cite{Annotated} \\
 & yang2023graphecho\cite{yang2023graphecho} \\
 & RadImageNet\cite{RadImageNet} \\
 & CuRIOUS2022\cite{Xiao2017RESECT,Behboodi2022RESECTSEG} \\ \midrule
\multirow{7}{*}{\rev{\textbf{Pathology}}} & PMC-VQA\cite{zhang2023pmc}\\
 & PathMMU\cite{PathMMU} \\
 & \rev{BCNB\cite{BCNB}} \\
 & \rev{lc25000\cite{lc25000}} \\
 & \rev{CRC100k\cite{CRC100k}} \\
 & \rev{Histopathologic Cancer Detection\cite{His}} \\
 & \rev{DigestPath19\cite{DigestPath19}} \\ \midrule
\multirow{2}{*}{\textbf{Clinical Photo}} & yu2025medframeqa\cite{yu2025medframeqa} \\
 & SMMILE\cite{smmile} \\ \bottomrule[1.5pt]
\end{tabular}
}
\vspace{-0.5em}
\end{table}

\begin{table}[H]
\centering
\setlength{\tabcolsep}{2pt}
\caption{Task Definitions of \bench.}
\label{tab:Task_Definition}
\resizebox{0.98\textwidth}{!}{
\begin{tabular}{@{}l|lcll@{}}
\toprule[1.5pt]
\textbf{Format} & \textbf{Task} & \textbf{\# Samples} & \textbf{Description} & \textbf{Subtask} \\
\midrule

\multirow{4}{*}{\makecell[l]{VQA}}
& Multiple choice & 300 & Select the correct answer from multiple options. & --- \\
\cmidrule(l){2-5}
& Blank filling & 300 & Fill in the missing keyword or phrase. & --- \\
\cmidrule(l){2-5}
& Report generation & 200 & Generate a detailed descriptive summary or report. & --- \\
\cmidrule(l){2-5}
& Question answering & 300 & Answer questions about the medical image in free-text form. & --- \\
\midrule

\multirow{14}{*}{\makecell[l]{Image\\Editing}}
& \multirow{4}{*}{\makecell[l]{Style transfer}}
& \multirow{4}{*}{900}
& \multirow{4}{*}{\makecell[l]{Transform a medical image into another\\specified visual or technical style.}}
& Programmatic style \\
& & & & Radiographic style \\
& & & & Dye style \\
& & & & General style \\
\cmidrule(l){2-5}

& \multirow{4}{*}{\makecell[l]{Artifact removal}}
& \multirow{4}{*}{599}
& \multirow{4}{*}{\makecell[l]{Remove scanning artifacts from medical\\images to improve image quality.}}
& Ring artifact \\
& & & & Motion artifact \\
& & & & Truncation artifact \\
& & & & Zipper artifact \\
\cmidrule(l){2-5}

& \multirow{3}{*}{\makecell[l]{Noise reconstruction}}
& \multirow{3}{*}{614}
& \multirow{3}{*}{\makecell[l]{Remove random noise from medical images\\and recover a clear original image.}}
& Gaussian noise \\
& & & & Salt-and-pepper noise \\
& & & & Poisson noise \\
\cmidrule(l){2-5}

& Resolution editing & 838 & Reconstruct the resolution of medical images. & Low-to-high resolution \\
\cmidrule(l){2-5}
& Contrast enhancement & 614 & Adjust the contrast of medical images. & Contrast enhancement \\
\cmidrule(l){2-5}
& Anatomical annotation & 307 & Add textual annotations of anatomical structures. & Region-based annotations \\
\midrule

\multirow{13}{*}{\makecell[l]{Multimodal\\Generation}}
& \multirow{3}{*}{\makecell[l]{Instruction editing}}
& \multirow{3}{*}{400}
& \multirow{3}{*}{\makecell[l]{Perform localized edits on specific regions\\of a medical image following natural-language instructions.}}
& Region-based enhancement \\
& & & & Organ-based annotation \& highlighting \\
& & & & Pathology simulation \\
\cmidrule(l){2-5}

& \multirow{3}{*}{\makecell[l]{organ-removal}}
& \multirow{3}{*}{86}
& \multirow{3}{*}{\makecell[l]{Remove specified organs or lesions from medical\\images and plausibly inpaint the background.}}
& Masked removal \\
& & & & Noise removal \\
& & & & Anatomical structure removal \\
\cmidrule(l){2-5}

& \multirow{2}{*}{\makecell[l]{organ-reconstruction}}
& \multirow{2}{*}{400}
& \multirow{2}{*}{\makecell[l]{Reconstruct missing or occluded organ structures\\from incomplete medical images.}}
& Masked reconstruction \\
& & & & Organ \& anatomical reconstruction \\
\cmidrule(l){2-5}

& \multirow{2}{*}{\makecell[l]{3D to 2D projection}}
& \multirow{2}{*}{400}
& \multirow{2}{*}{\makecell[l]{Extract and generate 2D cross-sectional slices\\from 3D medical volumetric data at specified positions.}}
& Grayscale slice \\
& & & & Colored slice \\
\cmidrule(l){2-5}

& \multirow{3}{*}{\makecell[l]{2D-to-3D reconstruction}}
& \multirow{3}{*}{164}
& \multirow{3}{*}{\makecell[l]{Reconstruct complete 3D volumetric data\\from a series of 2D medical image slices.}}
& Grayscale 3D reconstruction \\
& & & & Colored 3D reconstruction \\
& & & & 3D segmentation mask \\
\bottomrule[1.5pt]

\end{tabular}
}
\end{table}
\vspace{-0.1cm}
\section{Detailed Evaluation Protocol}
\label{app:evaluation_protocol}
\subsection{\rev{Detailed Evaluation Metrics}}
\label{app:metrics}

\rev{To compare outputs consistently, we use a common three-tier protocol across VQA, image editing, and multimodal generation. Let the input be an image $X_{\mathrm{in}}$ and instruction $T_{\mathrm{in}}$, and let a system produce image $X_{\mathrm{out}}$ and text $T_{\mathrm{out}}$. When an output type is applicable, it is compared with the reference $X_{\mathrm{gt}}$ or $T_{\mathrm{gt}}$ reviewed by clinical experts and GPT-4o.}

\subsection{\rev{Image Fidelity and Medical-Image Representation}}
\label{app:image_fidelity}

\rev{To quantify complementary aspects of reconstruction fidelity for image-generation and editing outputs, we systematically report full-image SSIM~\cite{MultiscaleSSIM}, PSNR, and LPIPS~\cite{lpips}. SSIM and PSNR quantify structural and pixel-wise fidelity, and LPIPS quantifies deep-feature perceptual distance. These measures are useful descriptions of reconstruction fidelity, but their global aggregation can miss small lesions, thin vessels, microcalcifications, subtle landmarks, and fine boundaries. Accordingly, they are reported as auxiliary measures rather than \mbox{clinical endpoints}.} \rev{To complement pixel- and perceptual-fidelity measures with a medical-image representation proxy, we additionally compute \textit{MedImageInsight Similarity} as the cosine similarity between frozen MedImageInsight embeddings of $X_{\mathrm{out}}$ and $X_{\mathrm{gt}}$~\cite{medimageinsight}. It supplies broader medical-image representation evidence than generic visual features, but it is neither lesion segmentation nor a substitute for clinical ground truth.} \rev{Dice, IoU, point/bounding-box accuracy, and surface-distance metrics are not reported. The fifteen benchmark tasks draw on heterogeneous source datasets and do not share a pixel-aligned spatial annotation ontology. Creating masks, boxes, points, or surfaces automatically would manufacture unsupported ground truth and change the benchmark. A separately expert-annotated localization subset is required before making spatial-localization claims.}

\subsection{\rev{Text Similarity and Answer Agreement}}
\label{app:text_similarity}

\rev{To quantify complementary forms of agreement with $T_{\mathrm{gt}}$ for text-producing tasks, we report BLEU~\cite{papineni2002bleu} and BERTScore~\cite{zhang2019bertscore} using PubMedBERT~\cite{PubMedBERT}. These measures capture lexical overlap and contextual semantic similarity, respectively.}

\rev{To provide an auditable closed-form answer record for the multiple-choice and blank-filling VQA tasks, we also report \textit{Exact Match (EM)}~\cite{squad2016}, which measures normalized exact agreement between the response and the reference answer. EM is not a measure of clinical quality for free text, open-ended answers, or reports, and it is not applicable to Gen. or Edit. outputs without a unique closed-form answer. We do not use keyword-style clinical-fact matching because it cannot reliably encode assertion status, anatomical site, laterality, severity, or \mbox{cross-modal clinical facts}.}

\rev{We do not include RadGraph-F1~\cite{delbrouck2022radgraph}, GREEN~\cite{green}, or RadCliQ~\cite{radcliq} in the benchmark-wide analysis because these metrics were developed and validated for chest-X-ray radiology-report evaluation. They are not defined for generated images and lack established validation for non-report VQA or the other five canonical modalities. A benchmark-wide aggregate would therefore exceed their supported scope.}

\subsection{\rev{Clinical Holistic Assessment}}
\label{app:clinical_holistic}

\rev{To apply a consistent clinical-holistic rubric while retaining task-specific requirements, we implement a checklist-guided VLM-as-a-Judge protocol~\cite{chen2024mllmasajudgeassessingmultimodalllmasajudge,zhang2025llmevalmed} using Lingshu-32B~\cite{lingshu2025}. The evaluation retains five shared dimensions: AA, CFA, and IC apply across all formats, while CMC and HOC apply only to contextual multimodal generation. Each canonical task supplies explicit core and secondary requirements as follows:}
\begin{equation}
\label{eq:checklist_definition}
\mathcal{K}_t=\mathcal{K}_{\mathrm{shared}}\cup\mathcal{K}_{\mathrm{task}}(t).
\end{equation}
\rev{For every requirement, the judge records whether it is supported by the available evidence and returns scores for each applicable dimension with brief evidence and an independent raw overall score. This keeps task criteria explicit while preserving one common metric family.}

\rev{Task-specific checklists operationalize clinically meaningful requirements within the applicable dimensions without adding top-level metrics. For VQA, they assess direct answers, image-supported anatomical and clinical claims, and omissions or inventions; for image editing, visible completion, preservation of patient-specific anatomy and non-target content, and introduced structures or artifacts; and for multimodal generation, requested anatomy and clinical state, image--text coherence, clinically important omissions, and unsupported findings.}

\rev{To distinguish agreement with a target reviewed by clinical experts and GPT-4o from input-based plausibility, we evaluate two complementary evidence settings using the same checklist. In the \textit{reference-aware} setting, the candidate is compared with the available image or expert- and GPT-4o-reviewed answer; in the \textit{reference-free} setting, the input image/instruction and candidate output are assessed without revealing the reference. Each applicable dimension and the holistic assessment use a 1--5 severity scale: 5, complete support; 4, substantively correct with minor issues; 3, partial completion or uncertainty; 2, a major core failure; and 1, irrelevant, critically incorrect, or unusable content. Lingshu-32B then assigns an independent raw holistic score, neither averaged across the applicable dimensions nor rescaled.}

\rev{To preserve the distinction between the two evidence settings, the main results report the raw holistic score separately for the two views, whereas checklist judgments and evidence support error analysis. Table~\ref{tab:researcher_audit_agreement} reports agreement between Lingshu-32B~\cite{lingshu2025} and ratings from three clinicians.}

\rev{\textbf{Medical-expert Audit Sampling.}}

\rev{To estimate judge--expert agreement across benchmark formats, we use the same 1,000 stratified cases and the same three clinicians described in Appendix~\ref{app:review}. Each clinician specializes in radiology, ultrasonography, or pathology and has at least three years of clinical experience. Lingshu-32B~\cite{lingshu2025} and the clinicians apply the shared overall 1--5 scale and five shared dimensions, with AA, CFA, and IC applying across formats and CMC and HOC applying only to contextual multimodal generation.}

\subsection{\rev{Aggregation and Error Analysis}}
\label{app:aggregation_error}

\rev{To identify clinical failure modes, we use clinical judge outputs to analyze clinical-finding accuracy and instruction compliance across formats and hallucination/omission control for contextual multimodal-generation cases. To quantify uncertainty and control multiplicity in the behavioral pairing analysis, Table~\ref{tab:entanglement_control} reports bootstrap 95\% confidence intervals and paired Wilcoxon signed-rank tests with Holm correction.}

\revisionon

\subsection{Detailed Benchmark-Construction Expert Review}
\label{app:review}

\textbf{Review Protocol and Sampling.}
\rev{We generated 9,000 candidate records and drew a stratified random sample of 1,000 records by task, format, and modality, covering every reported task and modality. Three clinical experts specializing in radiology, ultrasonography, and pathology, each with at least three years of clinical experience, independently reviewed all 1,000 sampled records.}

\textbf{Independent Blinded Review and Adjudication.} \rev{Blinded to the GPT-4o screening result, experts reviewed question reasonableness, answer correctness, and multimodal relevance. Non-unanimous decisions were independently reviewed by a fourth senior clinical expert, whose decision to retain or reject was final. The same 1,000 cases and three experts underpin the Lingshu-32B clinical evaluation. The GPT-4o versus expert construction screening and Lingshu-32B versus expert evaluation comparisons are reported separately.}

\textbf{Agreement with GPT-4o.} \rev{Experts retained 768 of the 1,000 records, whereas GPT-4o retained 648. Both retained 564 records; experts alone retained 204, GPT-4o alone retained 84, and both excluded 148. Relative to the final expert decision, the raw agreement was 71.2\% and Cohen's $\kappa$ was 0.315. Across 10,000 nonparametric bootstrap resamples of the 1,000 paired decisions, the 95\% percentile confidence interval was 68.3\%--74.0\% for raw agreement and 0.253--0.375 for Cohen's $\kappa$.}

\textbf{Screening of the Remaining Candidates.} \rev{GPT-4o screened the remaining 8,000 candidates using the same settings, retaining 5,654 and excluding 2,346 low-quality records. Together with the 768 records retained by the final expert decision in the stratified sample, these records yielded the 6,422 released pairs reviewed by clinical experts and GPT-4o.}

\begin{figure}[!t]
\centering
\begin{minipage}[b]{0.475\textwidth}
\centering
\includegraphics[width=\linewidth]{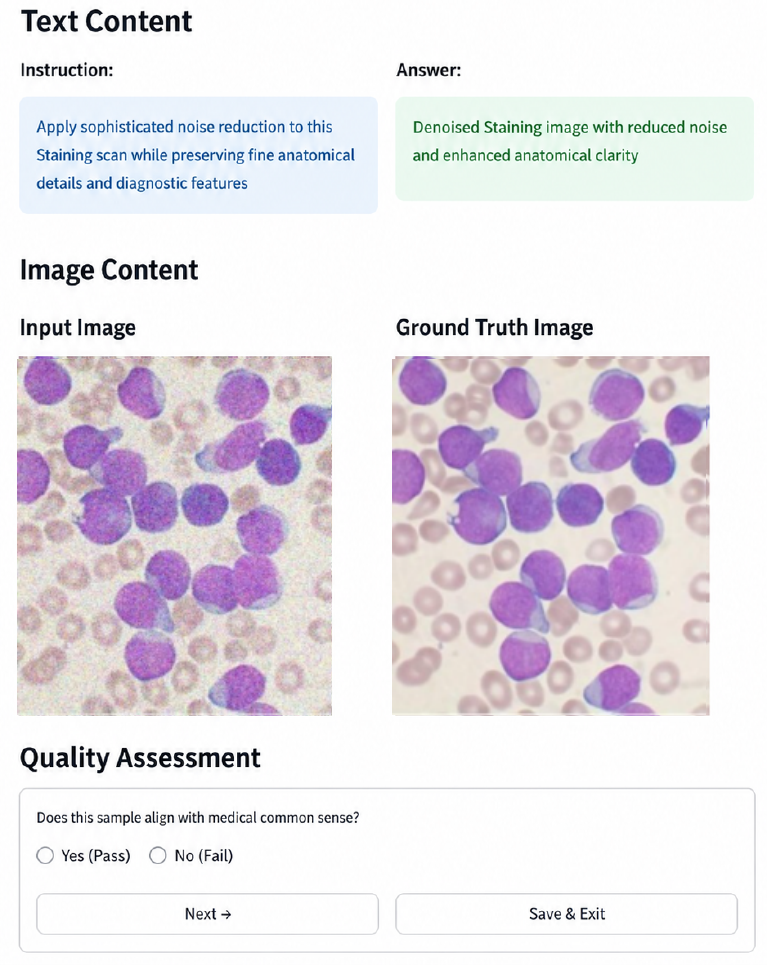}
\captionof{figure}{Annotation interface provided to support consistent expert review and filtering of candidate benchmark samples. The figure shows the case context, paired images, and quality-assessment controls available to reviewers.}
\label{fig:anno}
\end{minipage}\hfill
\begin{minipage}[b]{0.475\textwidth}
\begin{tcolorbox}[expertbox]
\vspace{-0.45em}
Thank you for contributing to this research effort! You are helping build a high-quality medical multimodal generation benchmark to evaluate and improve multimodal AI systems in clinical contexts.

\smallskip
Each sample consists of two image-text pairs:
\begin{itemize}\setlength{\itemsep}{1pt}\setlength{\topsep}{1pt}\setlength{\parskip}{0pt}
\item \textbf{Prompt pair}: an instruction (question or task) paired with an \emph{input medical image}, designed to challenge the AI model.
\item \textbf{Reference pair}: the expected \emph{answer text} paired with a \emph{ground truth image} that illustrates or validates the correct response.
\end{itemize}

\smallskip
When reviewing, please assess the sample along the following dimensions:
\begin{enumerate}\setlength{\itemsep}{1pt}\setlength{\topsep}{1pt}\setlength{\parskip}{0pt}
\item \textbf{Question Reasonableness}: Is the instruction clinically meaningful, unambiguous, and appropriate for the input image? Avoid questions that are trivial, misleading, or based on non-visible features.
\item \textbf{Answer Correctness}: Does the provided answer accurately reflect medical knowledge and align with the ground truth image? Ensure it is factually sound and specific (e.g., ``Pleural effusion, right side'' rather than ``There is fluid'').
\item \textbf{Multimodal Relevance}: Is there a clear, necessary connection between the instruction and the input image? The question should require visual interpretation — i.e., it cannot be answered from text alone.
\end{enumerate}
\vspace{-0.45em}
\end{tcolorbox}
\end{minipage}
\end{figure}

\textbf{Annotation Interface.} To support consistent expert screening, Figure~\ref{fig:anno} shows the annotation interface used to review and filter candidate data.

\ifnum\MarkedUpCopy=1
    \colorlet{expertbody}{\revisioncolor}
\else
    \colorlet{expertbody}{black}
\fi
\tcbset{
    expertbox/.style={
        colback=white,
        colframe=black,
        colbacktitle=black,
        coltitle=white,
        coltext=expertbody,
        fonttitle=\bfseries,
        fontupper=\footnotesize,
        title={Instructions for Clinical Experts},
        boxsep=3pt,
        arc=1.5mm, auto outer arc,
        left=6pt, right=6pt, top=3pt, bottom=3pt,
        boxrule=0.8pt,
        width=\linewidth
    }
}

\section{Detailed Instruction-Entanglement Analysis}
\label{app:ablation}
\revisionon





\subsection{Experimental Setup}
\label{app:entanglement_setup}
\revisionon
To determine whether contextual augmentation improves the image specificity of benchmark instructions, we systematically evaluate semantic relevance using the GME multimodal embedding model \cite{gme}. For each analyzed pair, we compare two types of instructions:
\begin{itemize}
\item \textbf{Original templates}: Instructions with placeholders (e.g., "Is [abnormality] present in [region]?")
\item \textbf{Augmented instructions}: Templates populated with structured clinical data (e.g., "Is pneumonia present in the right lower lobe?")
\end{itemize}
To provide external semantic baselines for this comparison, we additionally use 60 representative queries from six public medical VQA benchmarks (VQA-RAD, PathVQA, DrVD, SMMILE, OmniMedVQA, and MedFrameQA), with 10 queries per benchmark. Text and image embeddings are generated using GME-Qwen2-VL-2B-Instruct.

\newpage
\subsection{Similarity Computation}
\label{app:similarity_computation}
\revisionon
To quantify image--text semantic alignment in a common embedding space, we compute the cosine similarity between the embeddings of an image $X$ and text $T$ as follows:
\begin{equation}
\text{sim}(T, X) = \frac{\mathbf{e}_T^\top \mathbf{e}_X}{\|\mathbf{e}_T\| \|\mathbf{e}_X\|},
\end{equation}
where $\mathbf{e}_T = f_{\text{text}}(T)$ and $\mathbf{e}_X = f_{\text{image}}(X)$ denote the text and image embeddings generated by the GME model. For each sample $i$ with image $X_i$, we compute:
\begin{enumerate}
\item \textbf{Augmented instruction similarity}: $s_i^{\text{aug}} = \text{sim}(T_i^{\text{aug}}, X_i)$,
\item \textbf{Original template similarity}: $s_i^{\text{orig}} = \text{sim}(T_i^{\text{orig}}, X_i)$,
\item \textbf{Benchmark query similarities}: For each medical benchmark $k$ ($k=1,\dots,6$) with query set $Q_k$, the average similarity is:
\begin{equation}
\bar{s}_{i,k} = \frac{1}{|Q_k|} \sum_{q \in Q_k} \text{sim}(q, X_i).
\end{equation}
\item \textbf{Global benchmark average}: The mean similarity across all benchmarks:
\begin{equation}
\bar{s}_i^{\text{bench}} = \frac{1}{6} \sum_{k=1}^{6} \bar{s}_{i,k}.
\end{equation}
\end{enumerate}

\color{black}
\subsection{Evaluation Metrics}
\label{app:entanglement_metrics}
\revisionon
To assess whether augmentation improves alignment relative to generic medical queries, we systematically report two core metrics:
\begin{enumerate}
\item \textbf{Similarity Gain}: The improvement of augmented instructions over the benchmark average:
\begin{equation}
g_i = s_i^{\text{aug}} - \bar{s}_i^{\text{bench}}.
\end{equation}
Positive $g_i$ indicates superior semantic alignment compared to generic medical queries.
    
\item \textbf{Pass Rate}: The proportion of samples where augmented instructions exceed the benchmark average:
\begin{equation}
\text{Pass Rate} = \frac{1}{N} \sum_{i=1}^{N} \mathbf{1}\!\left[s_i^{\text{aug}} > \bar{s}_i^{\text{bench}}\right],
\end{equation}
where $\mathbf{1}[\cdot]$ is the indicator function and $N$ is the number of analyzed pairs.
\end{enumerate}
To measure the direct within-pair change produced by augmentation, we additionally compute the unnormalized similarity difference:
\begin{equation}
\Delta s_i = s_i^{\text{aug}} - s_i^{\text{orig}}.
\end{equation}
The mean $\Delta s_i$ across samples quantifies the mean change in similarity attributable to contextual augmentation. To characterize distributional shifts in semantic relevance rather than only average effects, we report the mean ($\mu$), median, and interquartile ranges of $g_i$ and $\Delta s_i$.

\subsection{Semantic Alignment of Contextual Instructions}
\label{app:semantic_alignment}
\revisionon
To connect construction-level semantic alignment with behavioral dependence on the intended pairing, Figure~\ref{fig:entanglement} reports the similarity distributions and Table~\ref{tab:entanglement_control} reports the pairing-sensitivity controls. Figure~\ref{fig:entanglement} shows positive similarity gains across all six benchmark comparisons and a rightward shift in the augmented-instruction distribution. The mean image--instruction similarity increases from 0.273 to 0.372, indicating stronger image-specific semantic alignment. Table~\ref{tab:entanglement_control} then complements this construction-level evidence with behavioral \mbox{pairing-sensitivity results}.

\subsection{\rev{Representative Examples}}
\label{app:representative_examples}
\rev{Paired cases compare qualitative successes and failures with medical images, task context, model output, and interpretation in one frame. Green text marks task-satisfying evidence; amber, qualified or partial evidence; red, substantive errors or unsupported claims. AA/CFA denote anatomical/clinical-finding accuracy; IC, instruction compliance; CMC, cross-modal consistency; HOC, hallucination/omission control. Overall is an independent holistic score; a dash indicates an unreported criterion.}

\rev{Case selection is diagnostic rather than frequency-representative: the stored outputs expose missing image outputs, replacement of case-specific anatomy, and confident grounding errors. They do not estimate failure prevalence or rank systems; instead, Figures~\ref{fig:cases_generation} and~\ref{fig:cases_editing} show how the shared dimensions support error analysis beyond aggregate scores. Reference-aware and reference-free rows are juxtaposed because a plausible output can still diverge from the reviewed patient-specific target. Reference metrics remain complementary evidence: neither high full-image similarity nor fluent text alone establishes task completion or clinical correctness. Interpretations therefore prioritize the requested operation, preservation of non-target anatomy, clinical findings, cross-modal support, and clinically relevant omissions. The closed- versus open-ended comparison appears in the main analysis (Section~\ref{sec:qualitative_openended}).}


\definecolor{caseblue}{RGB}{0,98,155}
\definecolor{casepositive}{RGB}{46,125,82}
\definecolor{casenegative}{RGB}{174,55,55}
\definecolor{caseamber}{RGB}{166,105,0}
\definecolor{casegray}{RGB}{247,248,249}
\ifnum\MarkedUpCopy=1
  \colorlet{casebody}{\revisioncolor}
\else
  \colorlet{casebody}{black}
\fi
\makeatletter
\setlength{\@dblfptop}{0pt}
\makeatother

\newcommand{\taskbanner}[1]{%
  \noindent\colorbox{caseblue}{%
    \parbox{\dimexpr\textwidth-2\fboxsep\relax}{%
      \centering\color{white}\sffamily\bfseries\small #1}}
}

\newcommand{\casecolumnbanner}[1]{%
  \noindent\colorbox{caseblue}{%
    \parbox{\dimexpr\columnwidth-2\fboxsep\relax}{%
      \centering\color{white}\sffamily\bfseries\scriptsize #1}}
}

\newtcolorbox{casecard}[3][]{
  enhanced,
  colback=white,
  colframe=#3,
  colbacktitle=#3,
  coltitle=white,
  coltext=casebody,
  fonttitle=\sffamily\bfseries\small,
  title={#2},
  boxrule=0.65pt,
  arc=1.2mm,
  left=5pt,
  right=5pt,
  top=6pt,
  bottom=6pt,
  boxsep=0pt,
  before skip=5pt,
  after skip=5pt,
  #1
}

\newtcolorbox{formatcontrastpanel}[2]{
  enhanced,
  colback=#2!4,
  colframe=#2,
  colbacktitle=#2,
  coltitle=white,
  coltext=casebody,
  fonttitle=\sffamily\bfseries\footnotesize,
  title={#1},
  boxrule=0.55pt,
  arc=0.8mm,
  left=4pt,
  right=4pt,
  top=3pt,
  bottom=3pt,
  boxsep=0pt
}

\newcommand{\caseimage}[3][2.45cm]{%
  \begin{minipage}[t]{0.315\linewidth}
    \centering
    \includegraphics[width=\linewidth,height=#1,keepaspectratio]{#2}\\[-0.25em]
    {\scriptsize #3}
  \end{minipage}%
}

\newcommand{\caseimageaccent}[4][2.45cm]{%
  \begin{minipage}[t]{0.315\linewidth}
    \centering
    \begingroup
    \setlength{\fboxsep}{1pt}\setlength{\fboxrule}{0.8pt}%
    \fcolorbox{#4}{white}{\includegraphics[width=0.92\linewidth,height=#1,keepaspectratio]{#2}}%
    \endgroup\\[-0.2em]
    {\scriptsize\textcolor{#4}{\bfseries #3}}
  \end{minipage}%
}
\newcommand{\caseimagegood}[3][2.45cm]{\caseimageaccent[#1]{#2}{#3~\ding{51}}{casepositive}}
\newcommand{\caseimagebad}[3][2.45cm]{\caseimageaccent[#1]{#2}{#3~\ding{55}}{casenegative}}

\newcommand{\singlecaseimage}[2]{%
  \begin{center}
    \includegraphics[width=0.90\linewidth,height=5.5cm,keepaspectratio]{#1}\\[-0.25em]
    {\scriptsize #2}
  \end{center}%
}

\newcommand{\casefield}[2]{%
  \par\smallskip\noindent\textbf{#1.} #2
}

\newcommand{\goodtext}[1]{\textcolor{casepositive!85!black}{\bfseries #1}}
\newcommand{\badtext}[1]{\textcolor{casenegative!90!black}{\bfseries #1}}
\newcommand{\partialtext}[1]{\textcolor{caseamber}{\bfseries #1}}
\newcommand{\scorehigh}[1]{\textcolor{casepositive!85!black}{\bfseries #1}}
\newcommand{\scoremid}[1]{\textcolor{caseamber}{\bfseries #1}}
\newcommand{\scorelow}[1]{\textcolor{casenegative!90!black}{\bfseries #1}}
\newcommand{\scorena}{\textcolor{black!45}{--}}

\newcommand{\generatedtext}[1]{%
  \begin{tcolorbox}[
    colback=casegray,
    colframe=black!16,
    coltext=casebody,
    boxrule=0.35pt,
    arc=0.6mm,
    left=3pt,right=3pt,top=2pt,bottom=2pt,
    boxsep=0pt,
    before skip=3pt,after skip=2pt]
    \textbf{Model-generated text.} #1
  \end{tcolorbox}%
}

\newcommand{\generatederror}[2]{%
  \begin{tcolorbox}[
    colback=casenegative!6,
    colframe=casenegative!75!black,
    coltext=casebody,
    boxrule=0.55pt,
    arc=0.6mm,
    left=3pt,right=3pt,top=2pt,bottom=2pt,
    boxsep=0pt,
    before skip=3pt,after skip=2pt]
    \badtext{#1.} #2
  \end{tcolorbox}%
}

\newcommand{\metricribbon}[1]{%
  \par\smallskip
  \noindent\colorbox{casegray}{%
    \parbox{\dimexpr\linewidth-2\fboxsep\relax}{%
      \footnotesize\textbf{Reference metrics:} #1}}
}

\newcommand{\casescoretable}[2]{%
  \par\smallskip
  {\footnotesize
  \setlength{\tabcolsep}{4.6pt}
  \renewcommand{\arraystretch}{1.12}
  \begin{center}
  \begin{tabular}{@{}lcccccc@{}}
    \toprule
    View & AA & CFA & IC & CMC & HOC & Overall \\
    \midrule
    w. GT   & #1 \\
    w.o. GT & #2 \\
    \bottomrule
  \end{tabular}
  \end{center}}
}

\FloatBarrier

\begin{figure*}[!t]
\centering
\taskbanner{Representative Cases --- Multimodal Generation}

\begin{casecard}{Positive \ding{51}\quad Anatomically coherent reconstruction}{casepositive}
{\small
\noindent\textbf{Model:} Gemini-2.5-flash-image \quad
\textbf{Subtask:} organ-reconstruction \quad
\textbf{Modality:} Ultrasound

\smallskip
\begin{minipage}[t]{0.45\linewidth}
  \vspace{0pt}
  \caseimage[3.5cm]{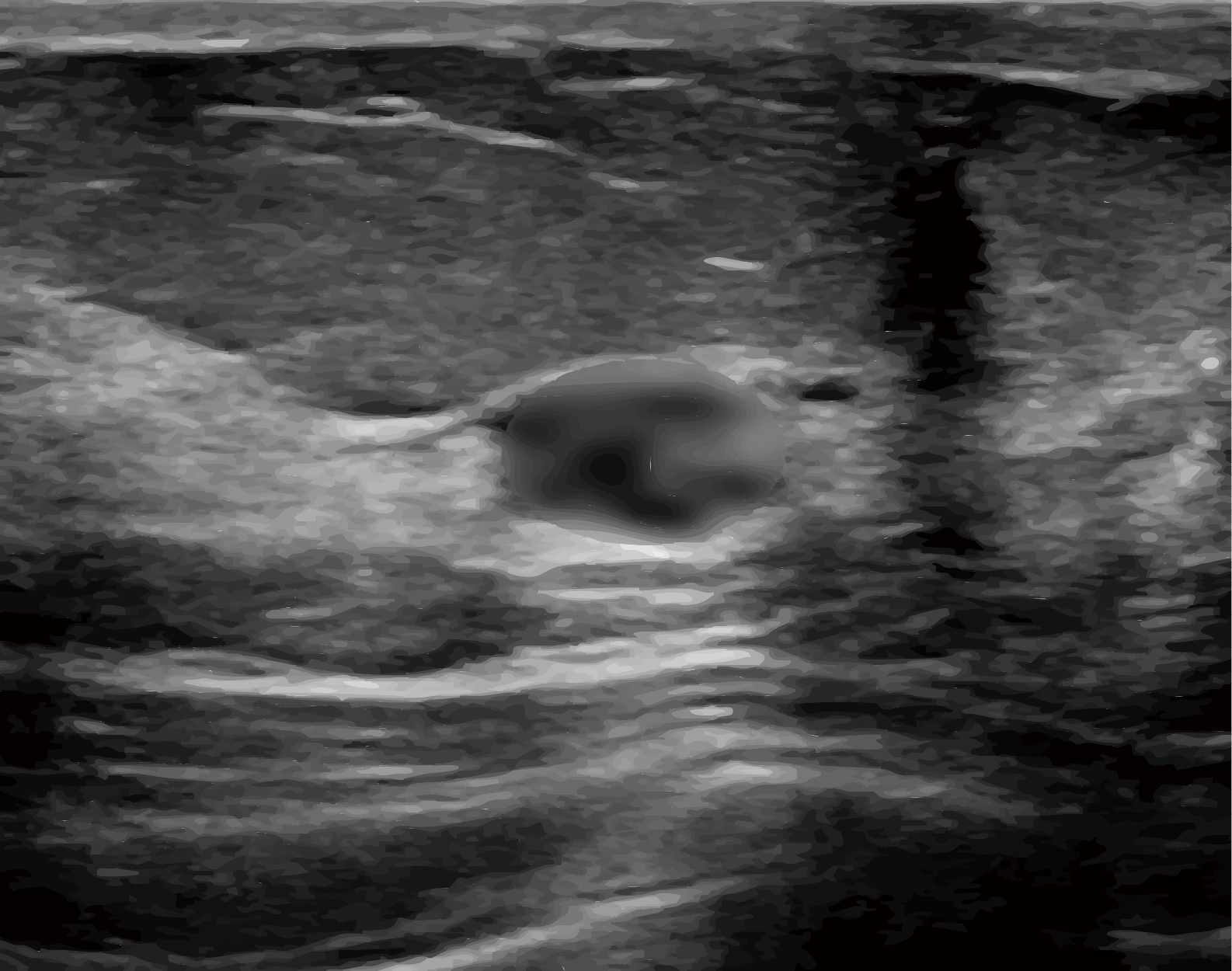}{(a) Input}
  \hfill
  \caseimage[3.5cm]{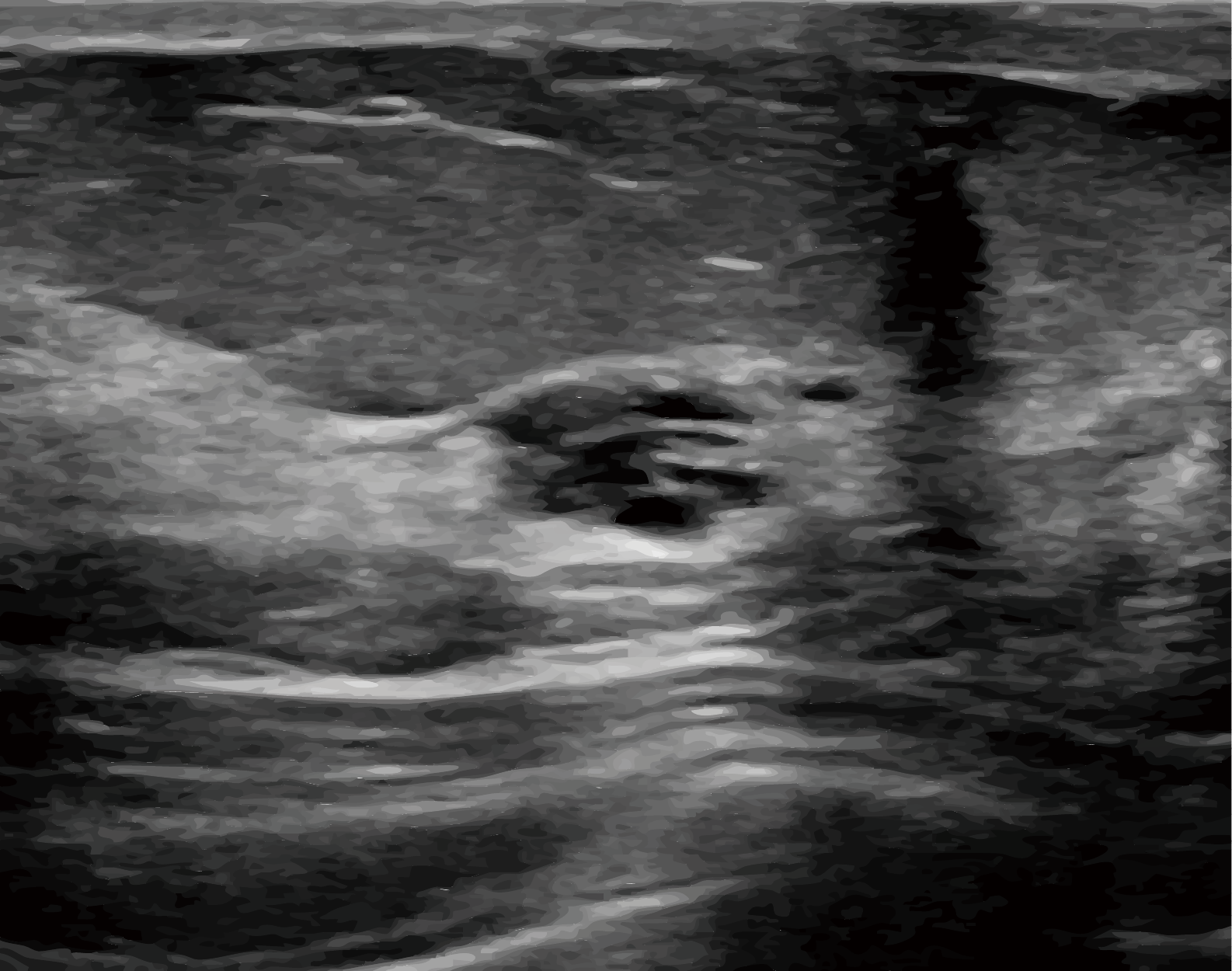}{(b) Ground truth}
  \hfill
  \caseimagegood[3.5cm]{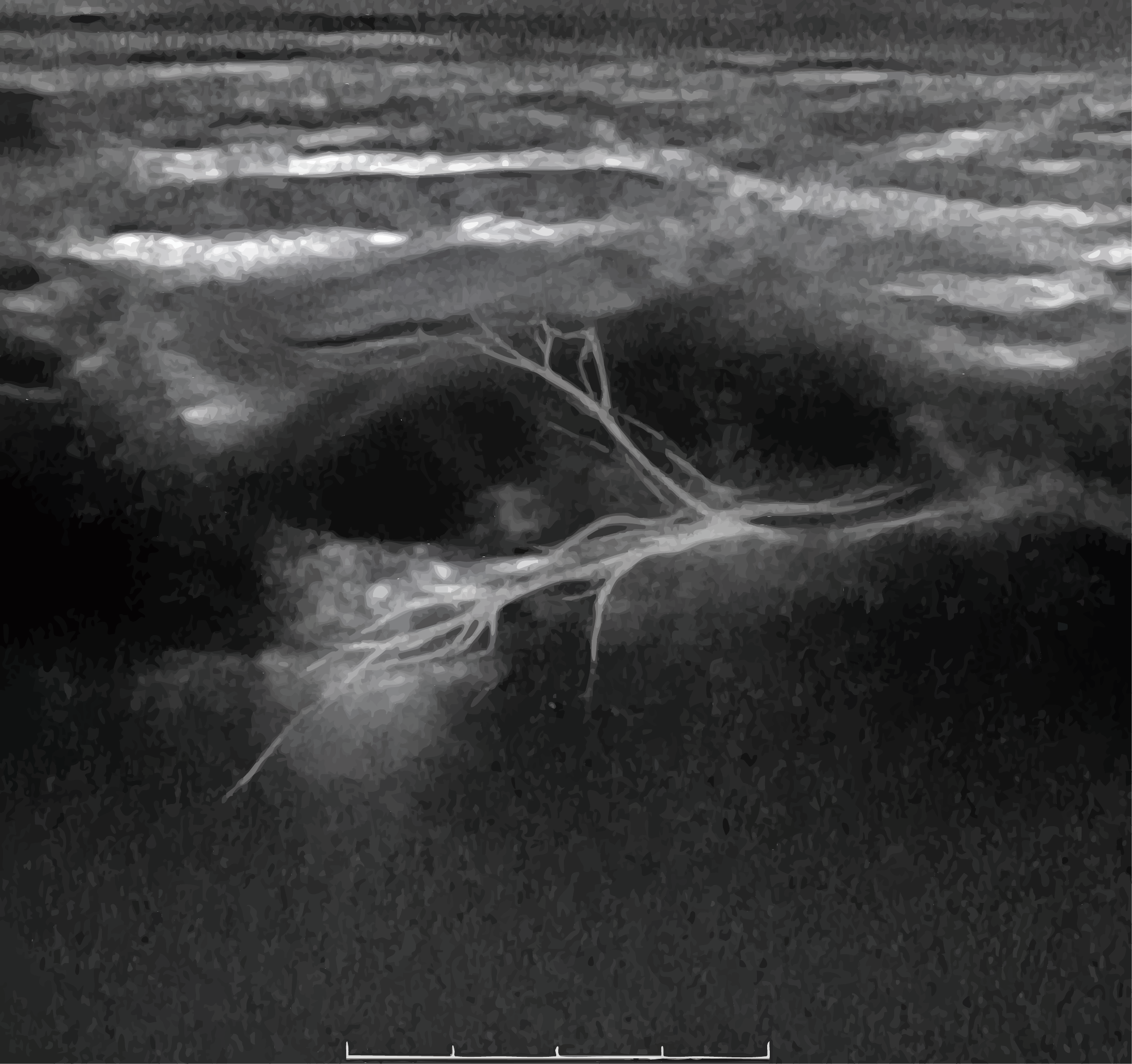}{(c) Model output}
\end{minipage}\hfill
\begin{minipage}[t]{0.525\linewidth}
  \vspace{0pt}
  \casefield{Instruction}{``Reconstruct the brachial plexus in this ultrasound image.''}
  \casefield{\goodtext{Why it performs well}}{The output \goodtext{reconstructs the missing brachial-plexus region}, while \goodtext{preserving surrounding tissue texture and anatomical continuity}. These properties support a favorable qualitative assessment.}
\end{minipage}

\metricribbon{LPIPS 0.25 \quad PSNR 29.41 \quad SSIM 0.43}
\casescoretable{\scorehigh{4} & \scorehigh{4} & \scorehigh{5} & \scorena & \scorena & \scorehigh{4}}{\scorehigh{4} & \scorehigh{4} & \scorehigh{5} & \scorena & \scorena & \scorehigh{4}}
}
\end{casecard}

\begin{casecard}{Negative \ding{55}\quad Image operation replaced by textual description}{casenegative}
{\small
\noindent\textbf{Model:} Qwen3-VL \& Seedream-4.0 (compositional framework) \quad
\textbf{Subtask:} organ-removal \quad
\textbf{Modality:} Medical Imaging

\smallskip
\begin{minipage}[t]{0.45\linewidth}
  \vspace{0pt}
  \caseimage[3.5cm]{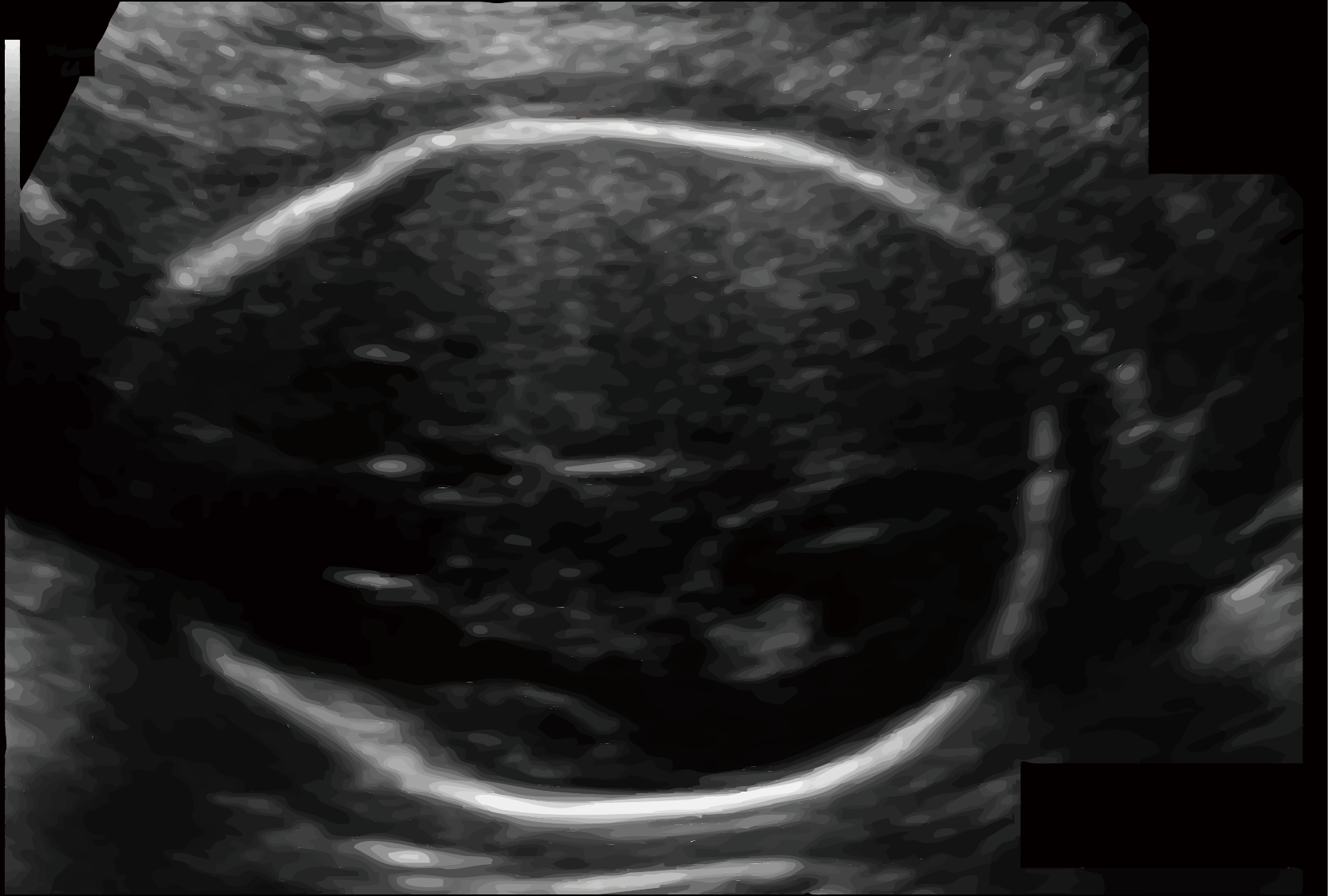}{(a) Input}
  \hfill
  \caseimage[3.5cm]{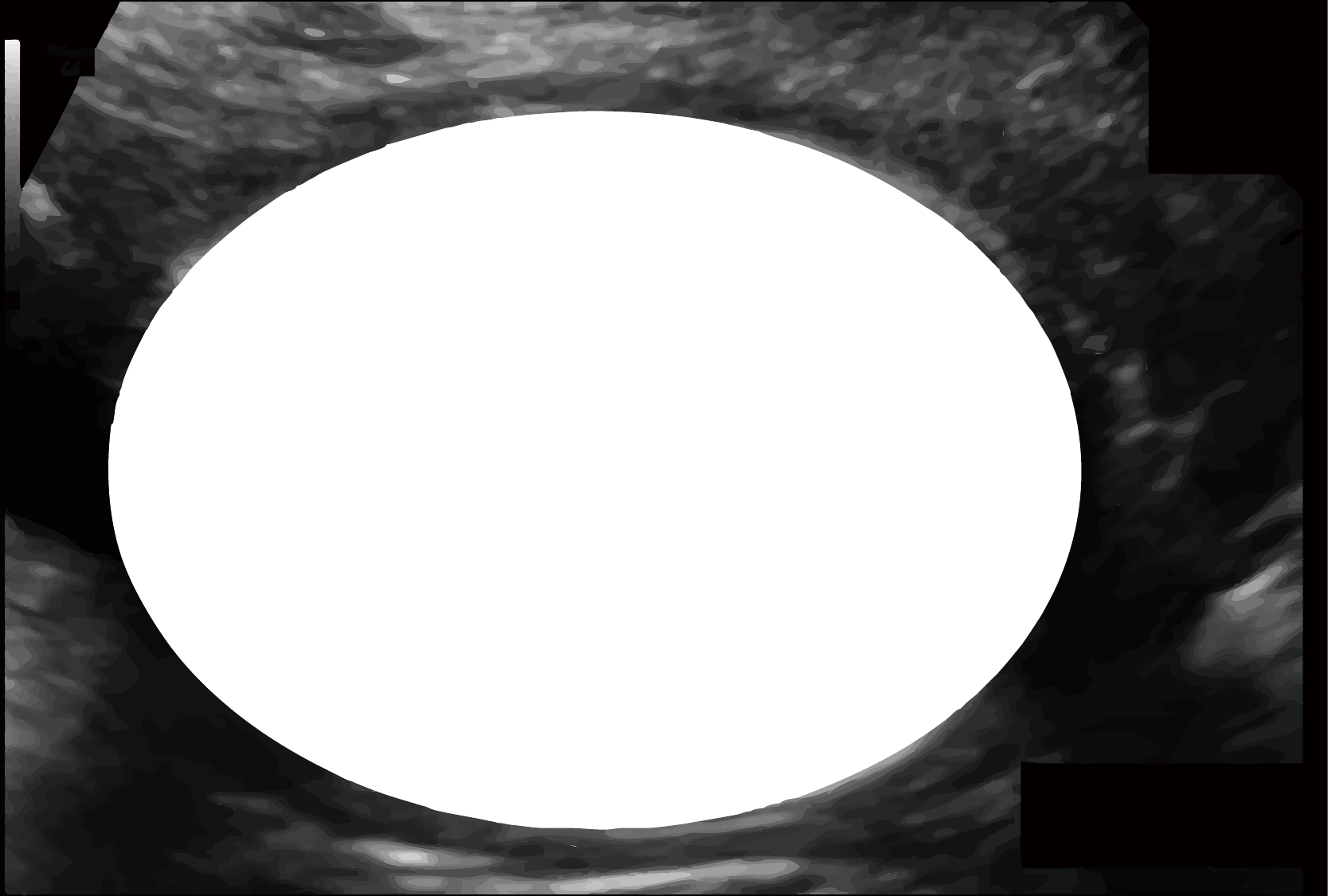}{(b) Ground truth}
  \hfill
  \caseimagebad[3.5cm]{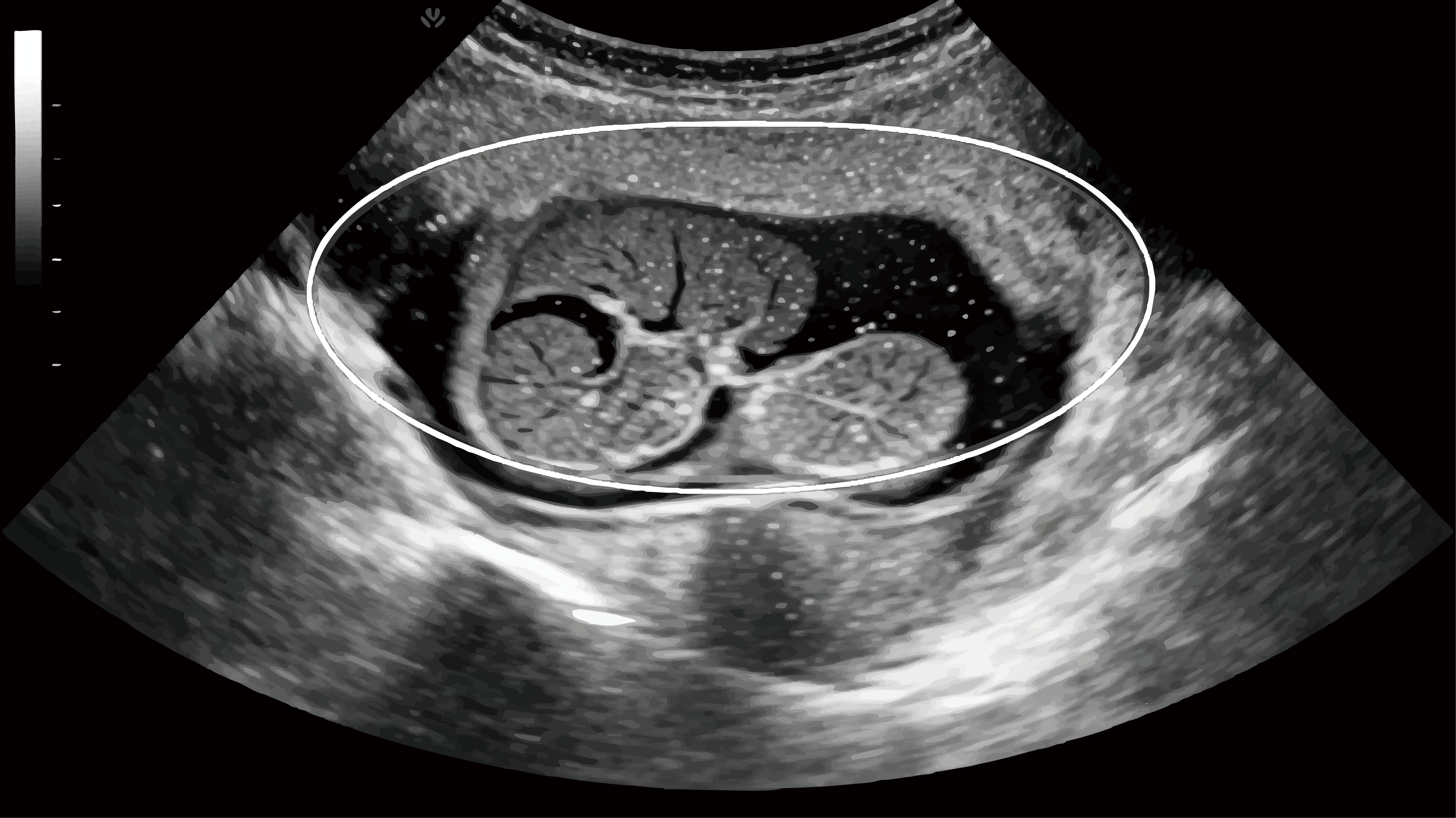}{(c) Model output}

  \casefield{Instruction}{``Remove the head circumference white from this medical image using fill technique.''}
\end{minipage}\hfill
\begin{minipage}[t]{0.525\linewidth}
  \vspace{0pt}
    \generatederror{\textcolor{red}{Wrong output modality.}}{%
    The model \textcolor{red}{returns a textual description instead of the requested edited image}. \ldots{}
    It \textcolor{red}{does not execute the requested fill/inpainting operation}; therefore,
    \textcolor{red}{the white head-circumference line is not removed}. \ldots{}%
    }
  \casefield{\badtext{Failure}}{The model \badtext{does not execute the requested fill operation}: it returns a textual explanation instead of a modified image. Its internally coherent narrative can appear plausible without the reference, but it does not compensate for the missing visual edit.}
\end{minipage}

\metricribbon{LPIPS 0.51 \quad PSNR 28.23 \quad SSIM 0.15 \quad BERTScore 0.92}
\casescoretable{\scorelow{1} & \scorelow{2} & \scoremid{3} & \scorelow{2} & \scorelow{1} & \scorelow{2}}{\scorelow{1} & \scorelow{2} & \scorehigh{4} & \scorelow{2} & \scorelow{1} & \scorelow{2}}
}
\end{casecard}

\begin{casecard}[top=3pt,bottom=3pt,before skip=4pt,after skip=3pt]{VQA Positive \ding{51}\quad Correct image-grounded identification}{casepositive}
{\footnotesize
\noindent\textbf{Model:} HuatuoGPT-Vision \quad \textbf{Subtask:} Question--Answer \quad \textbf{Modality:} Radiograph

\begin{minipage}[t]{0.22\linewidth}
  \vspace{2pt}\centering
  \includegraphics[width=0.78\linewidth,height=2.2cm,keepaspectratio]{figure/representative_examples/Positive_sample_vqa_input_image.pdf}\\[-0.3em]
  {\scriptsize Input image}
\end{minipage}\hfill
\begin{minipage}[t]{0.755\linewidth}
  \vspace{0pt}
  \casefield{Question}{``Which color represents right lung in this image?''}
  \casefield{Model output}{``The color representing the right lung in this image is \goodtext{red}.''}
  \casefield{\goodtext{Why it performs well}}{The answer is \goodtext{correct, concise, and directly grounded in the color-coded anatomy}, supporting a favorable qualitative assessment.}
\end{minipage}

\metricribbon{BERTScore 0.95}
\casescoretable{\scorehigh{5} & \scorehigh{5} & \scorehigh{5} & \scorena & \scorena & \scorehigh{5}}{\scorehigh{5} & \scorehigh{5} & \scorehigh{5} & \scorena & \scorena & \scorehigh{5}}
}
\end{casecard}

\caption{\rev{Positive and negative multimodal-generation cases together with a successful VQA example. Images, instructions, model behavior, reference metrics, and case ratings are aligned within each case. CMC and HOC are reported only for contextual multimodal-generation cases; a dash denotes that the criterion is not applicable.}}
\label{fig:cases_generation}
\end{figure*}

\begin{figure*}[!t]
\centering
\taskbanner{Representative Cases --- Image Editing}

\begin{casecard}[top=5pt,bottom=5pt,before skip=4pt,after skip=4pt]{Positive \ding{51}\quad Artifact removal with preserved anatomy}{casepositive}
{\small
\noindent\textbf{Model:} Qwen-Image-Edit \quad
\textbf{Subtask:} Artifact Removal \quad
\textbf{Modality:} X-ray

\smallskip
\begin{minipage}[t]{0.45\linewidth}
  \vspace{0pt}
  \caseimage[3.7cm]{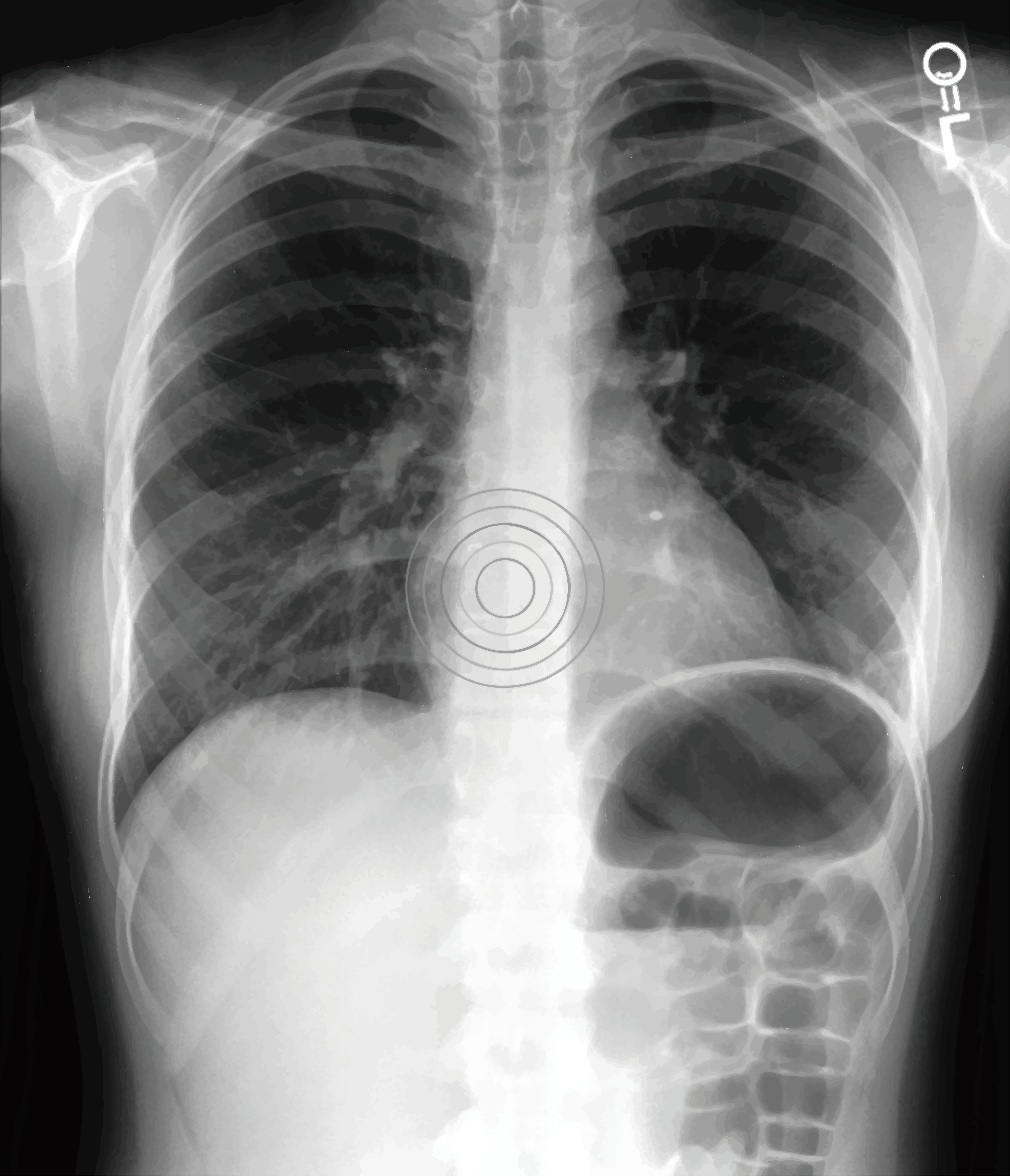}{(a) Input}
  \hfill
  \caseimage[3.7cm]{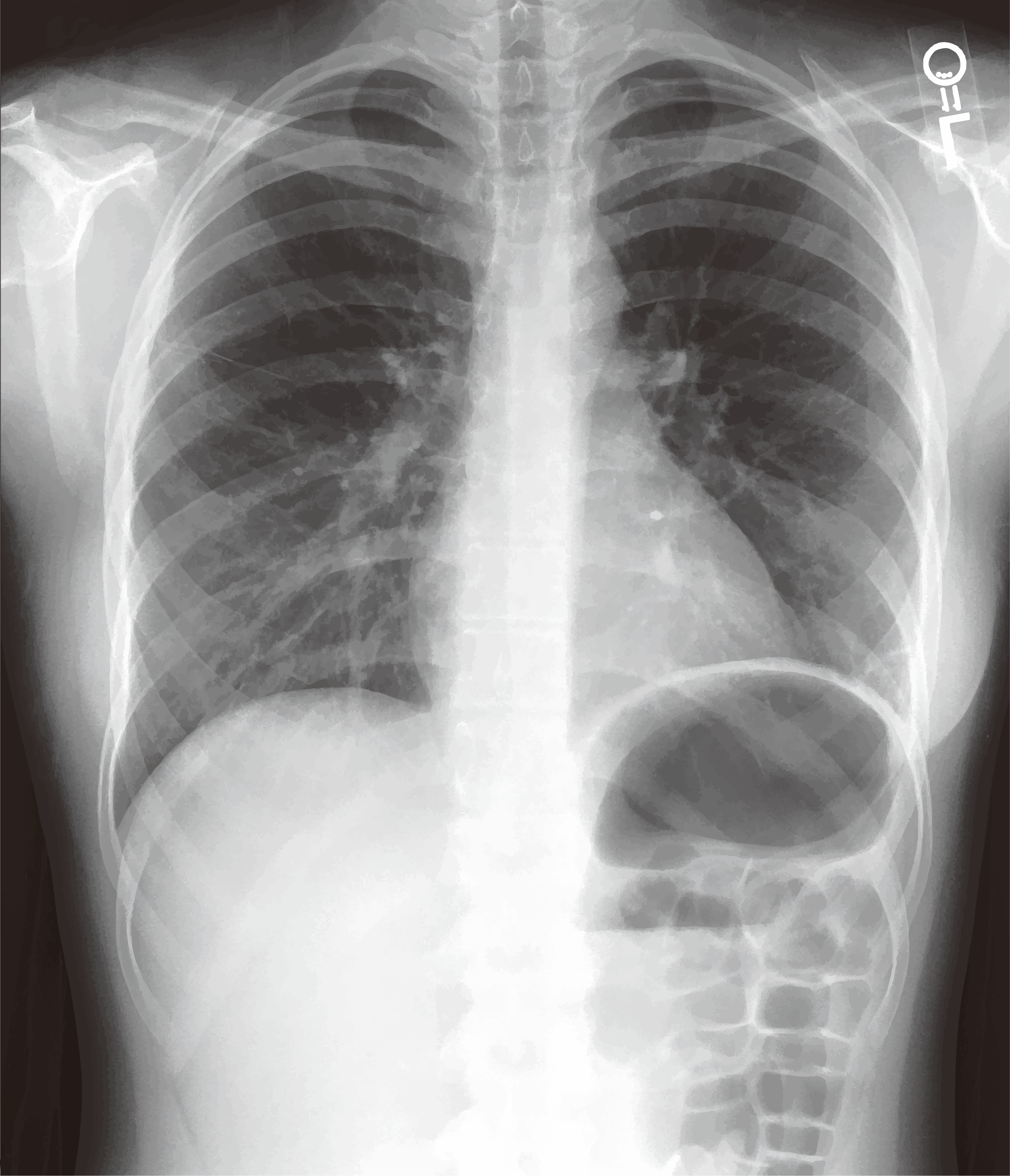}{(b) Ground truth}
  \hfill
  \caseimagegood[3.7cm]{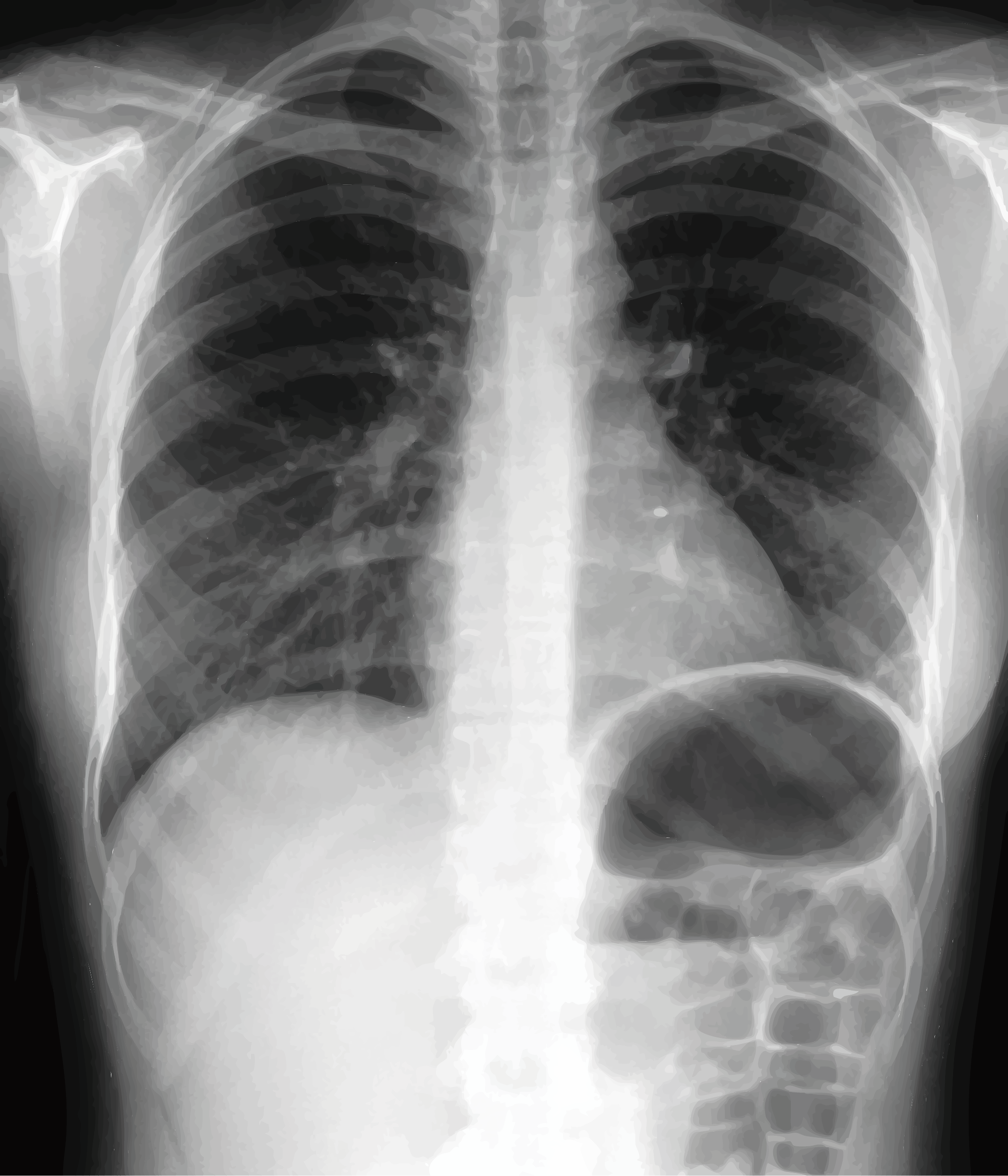}{(c) Model output}
\end{minipage}\hfill
\begin{minipage}[t]{0.525\linewidth}
  \vspace{0pt}
  \casefield{Instruction}{``Clean up this X-ray scan by eliminating the ring artifact while preserving anatomical structures and diagnostic information.''}
  \casefield{\goodtext{Why it performs well}}{The edit \goodtext{removes the ring artifact} while \goodtext{preserving the ribs, spine, lung fields, and diaphragm}. This supports a favorable qualitative assessment. \partialtext{The reference-aware view is less favorable because the output differs from the exact expert target despite being a plausible edit.}}
\end{minipage}

\metricribbon{LPIPS 0.06 \quad PSNR 29.65 \quad SSIM 0.78}
\vspace{-0.7em}
\casescoretable{\scorelow{1} & \scorelow{2} & \scorehigh{4} & \scorena & \scorena & \scoremid{3}}{\scorehigh{5} & \scorehigh{5} & \scorehigh{5} & \scorena & \scorena & \scorehigh{5}}
}
\end{casecard}

\begin{casecard}[top=5pt,bottom=5pt,before skip=4pt,after skip=4pt]{Negative \ding{55}\quad Synthetic replacement instead of contrast enhancement}{casenegative}
{\small
\noindent\textbf{Model:} Qwen3-VL \& GPT-image-1 \quad
\textbf{Subtask:} Contrast Enhancement \quad
\textbf{Modality:} Staining

\smallskip
\begin{minipage}[t]{0.45\linewidth}
  \vspace{0pt}
  \caseimage[3.7cm]{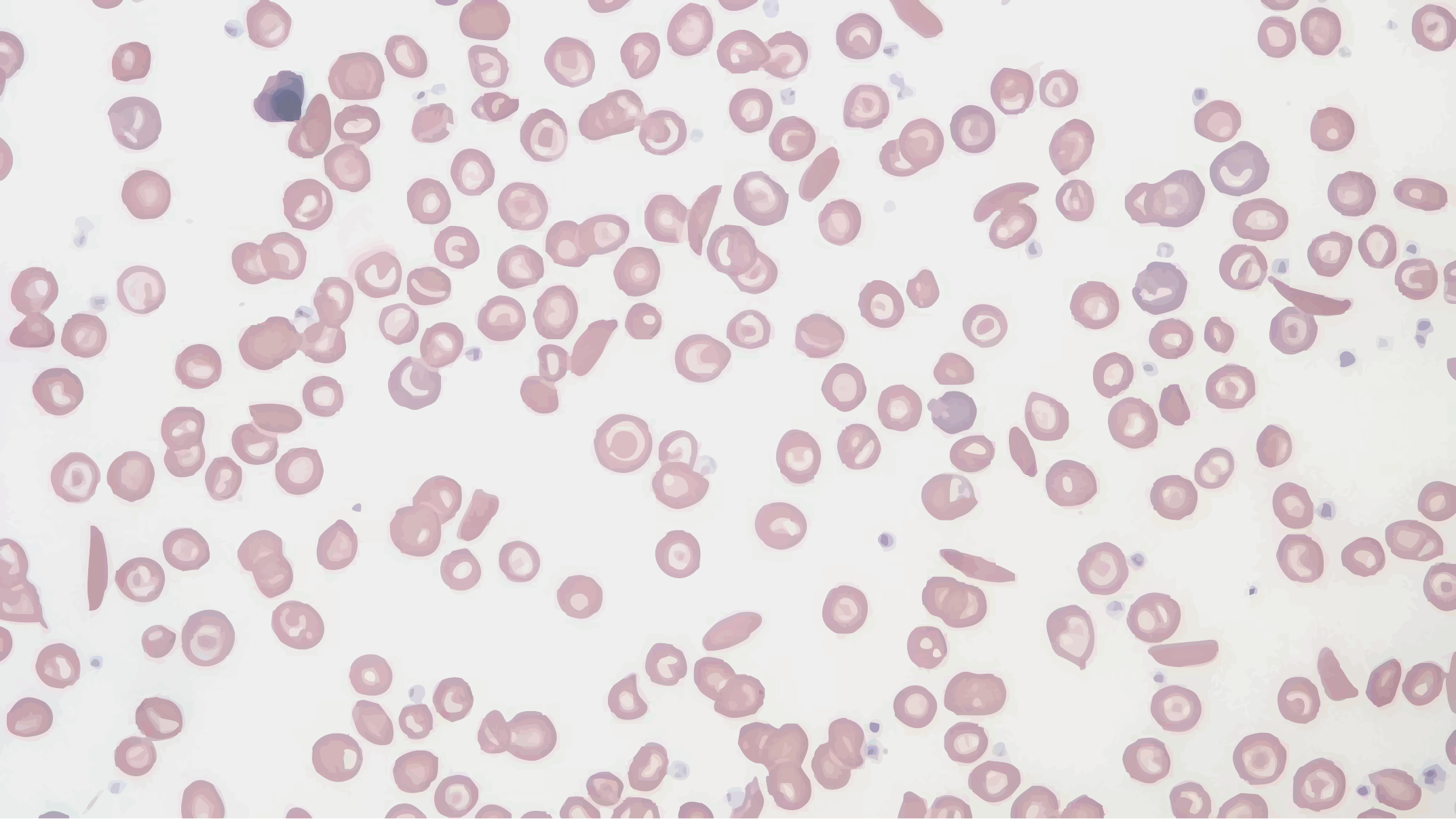}{(a) Input}
  \hfill
  \caseimage[3.7cm]{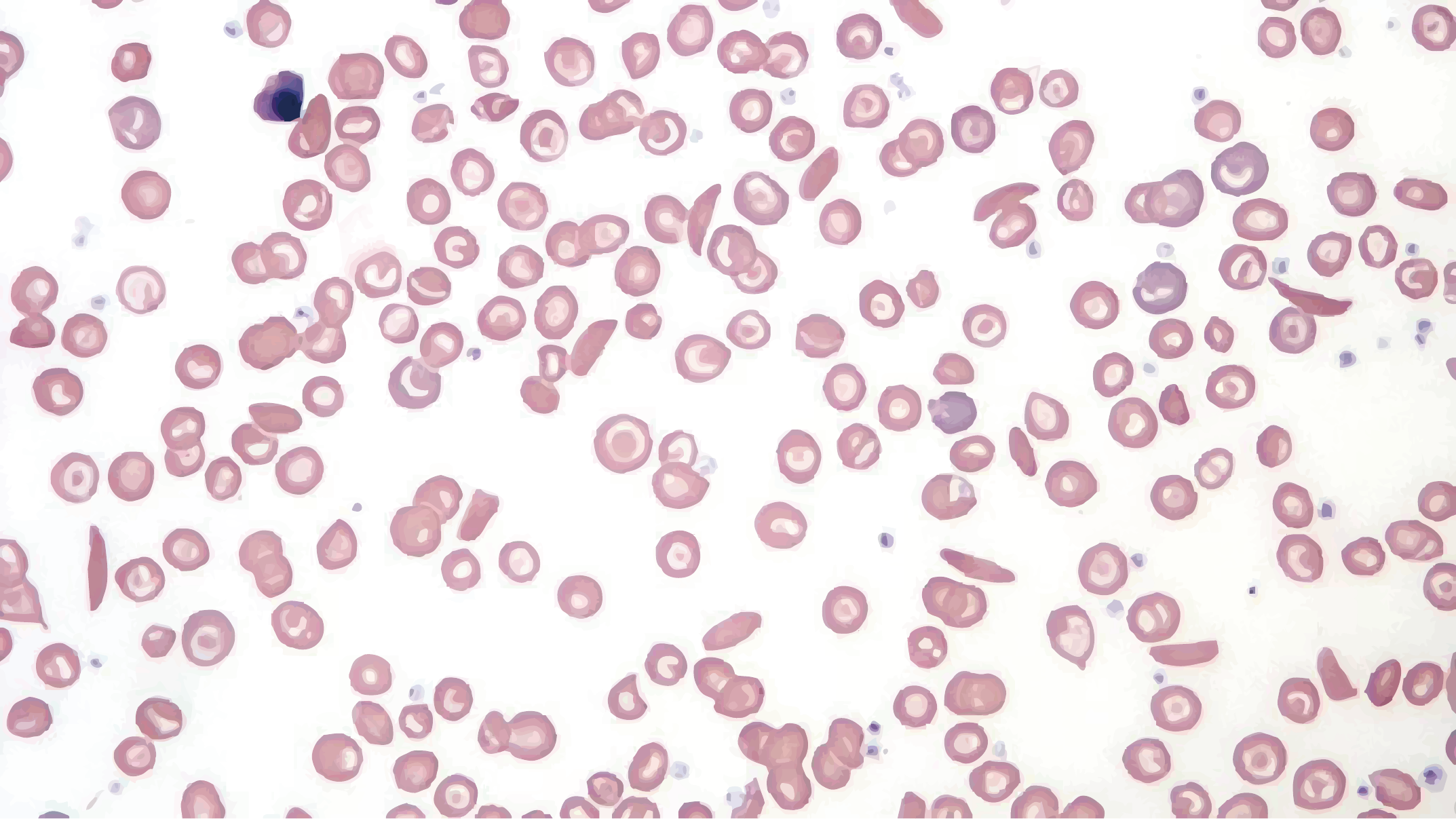}{(b) Ground truth}
  \hfill
  \caseimagebad[3.7cm]{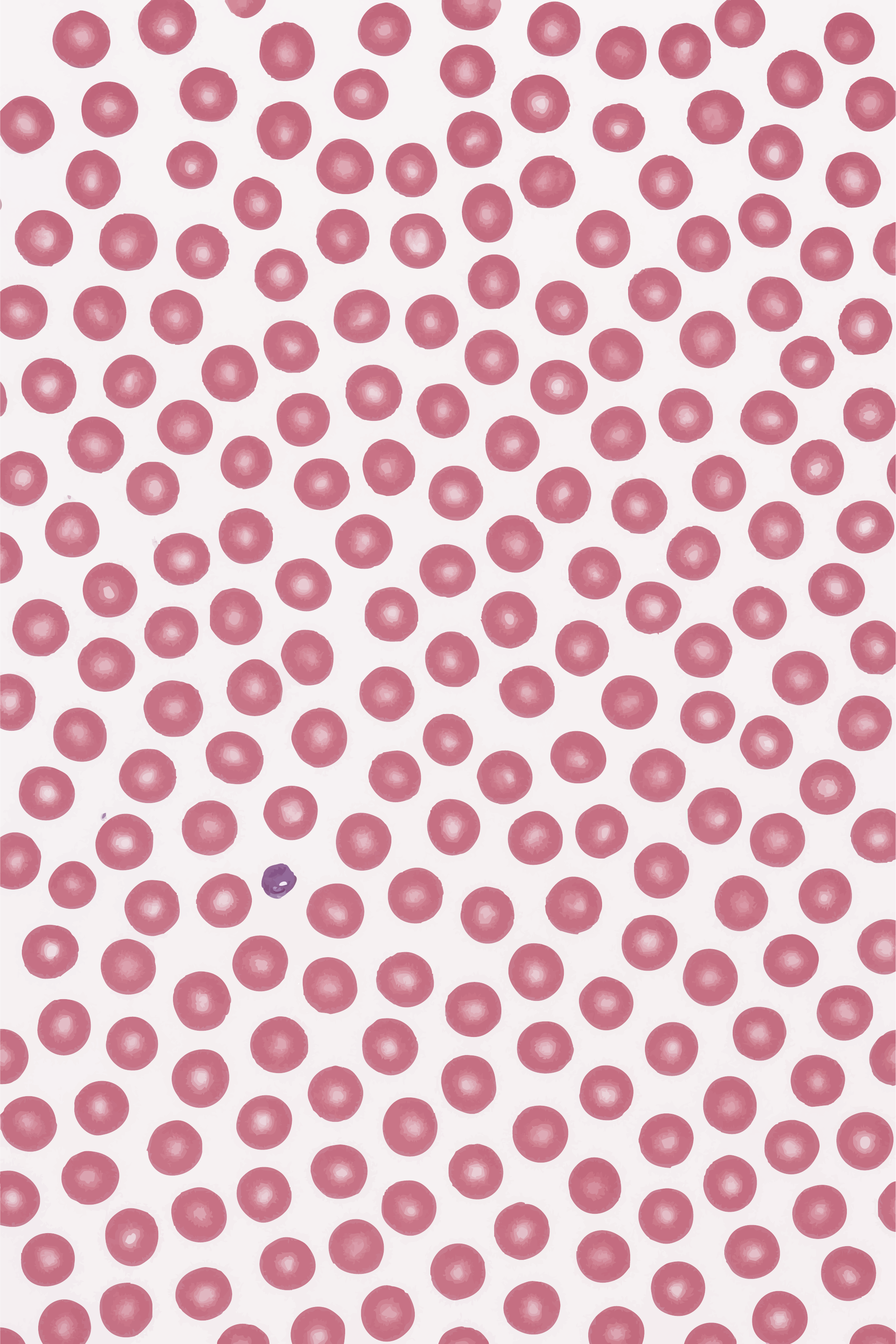}{(c) Model output}

  \casefield{Instruction}{``Enhance the Staining image contrast.''}
\end{minipage}\hfill
\begin{minipage}[t]{0.525\linewidth}
  \vspace{0pt}
    \generatederror{Output--image mismatch}{%
    The provided image is a stained peripheral blood smear showing red blood cells (RBCs) with varying morphology. \ldots{}
    The current contrast level is relatively low. \ldots{}
    \badtext{Enhancing the contrast will improve visualization of RBC shape, size variation (anisocytosis), and potential inclusion bodies or staining artifacts}. \ldots{}%
    }
  \casefield{\badtext{Failure}}{Instead of enhancing the original smear, the model \badtext{replaces it with an artificial synthetic pattern}, losing \badtext{case-specific cellular morphology and natural stain appearance}.}
\end{minipage}

\metricribbon{LPIPS 0.65 \quad PSNR 28.92 \quad SSIM 0.24}
\vspace{-0.7em}
\casescoretable{\scorelow{2} & \scoremid{3} & \scorehigh{4} & \scorena & \scorena & \scorelow{2}}{\scorelow{2} & \scorelow{1} & \scorelow{2} & \scorena & \scorena & \scorelow{2}}
}
\end{casecard}

\begin{casecard}[top=2pt,bottom=2pt,before skip=3pt,after skip=2pt]{VQA Negative \ding{55}\quad Confident anatomy-grounding error}{casenegative}
{\footnotesize
\noindent\textbf{Model:} Qwen3-VL \quad \textbf{Subtask:} Blank-Filling \quad \textbf{Modality:} MRI

\begin{minipage}[t]{0.22\linewidth}
  \vspace{2pt}\centering
  \includegraphics[width=0.82\linewidth,height=2.3cm,keepaspectratio]{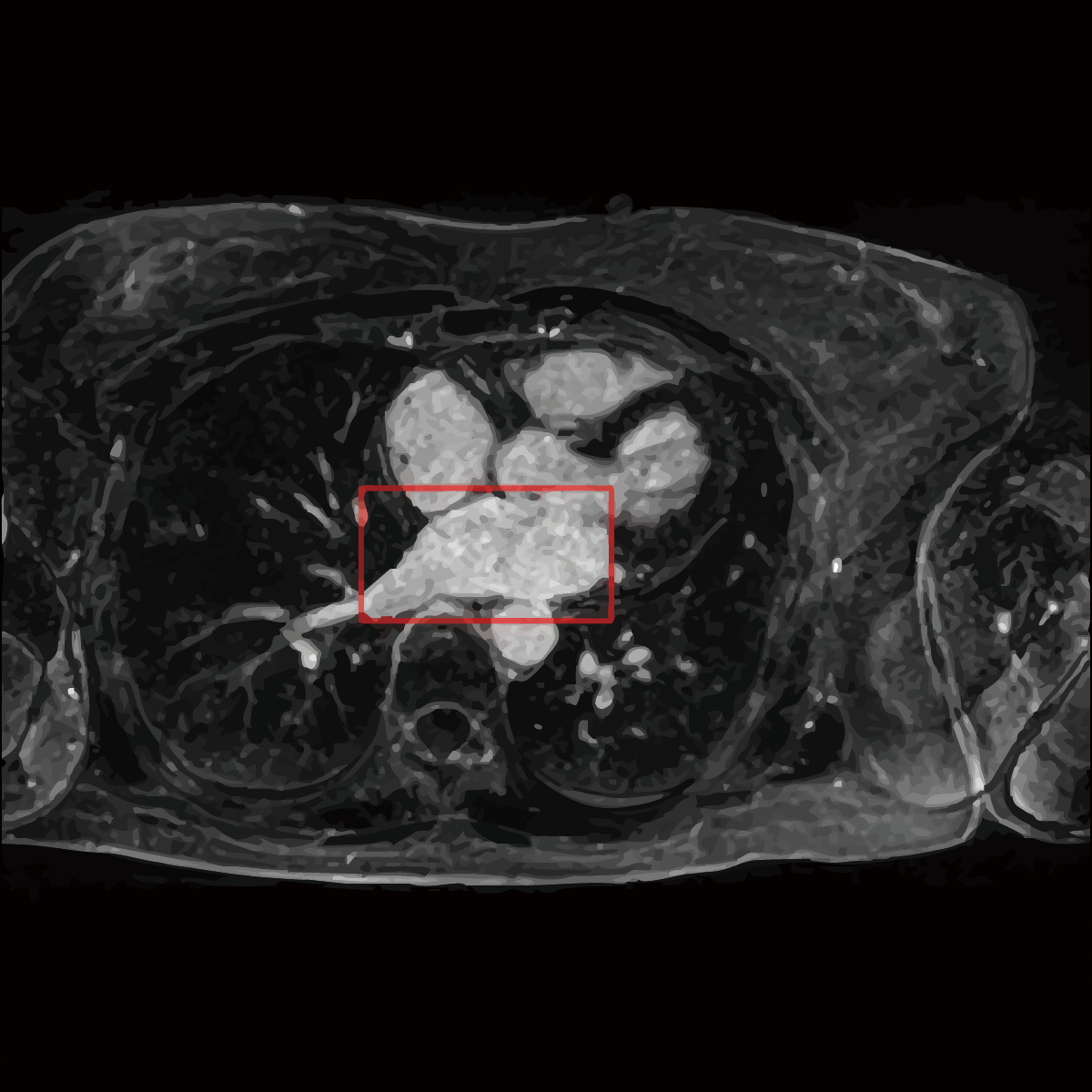}\\[-0.3em]
  {\scriptsize Input image}
\end{minipage}\hfill
\begin{minipage}[t]{0.755\linewidth}
  \vspace{0pt}
  \casefield{Question}{``The organ inside the red box in this image is the \_\_\_\_.''}
  \casefield{Model output}{``The organ inside the red box in this image is the \badtext{liver}. \badtext{The image appears to be a cross-sectional MRI scan of the abdomen}, and the structure enclosed in the red box \badtext{corresponds anatomically to the liver}, which is located in the \badtext{right upper quadrant of the abdominal cavity}. The liver's characteristic size, shape, and position such as \badtext{diaphragm superiorly and the stomach/gallbladder anteriorly} support this identification.''}
  \casefield{\badtext{Failure}}{The marked structure is \badtext{cardiac, not hepatic}. The fluent rationale preserves superficial textual coherence, but the anatomy-grounded assessment remains unfavorable because the central anatomy is wrong.}
\end{minipage}

\metricribbon{BERTScore 0.90}
\vspace{-0.7em}
\casescoretable{\scorelow{1} & \scorelow{2} & \scorehigh{4} & \scorena & \scorena & \scorelow{2}}{\scorelow{1} & \scorehigh{4} & \scorehigh{5} & \scorena & \scorena & \scoremid{3}}
}
\end{casecard}

\caption{\rev{Positive and negative image-editing cases together with a failed VQA example. The failures illustrate replacement of case-specific morphology and confident but incorrect anatomy grounding. CMC and HOC are not applicable to these cases.}}
\label{fig:cases_editing}
\label{fig:cases_vqa}
\end{figure*}

\FloatBarrier
\color{black}

\iftrue
\revisionon

\revisionon

\clearpage
\onecolumn
\makeatletter
\setlength{\@fptop}{0pt}
\setlength{\@fpsep}{6pt}
\setlength{\@fpbot}{0pt}
\makeatother

\ifnum\MarkedUpCopy=1
    \colorlet{promptbody}{\revisioncolor}
\else
    \colorlet{promptbody}{black}
\fi
\newtcolorbox{promptbox}[1]{
    colback=white,            
    colframe=black,           
    coltext=promptbody,
    fonttitle=\bfseries,      
    fontupper=\small,         
    title=#1,                 
    boxsep=4pt,               
    arc=1.5mm, auto outer arc,                
    left=8pt, right=8pt, top=4pt, bottom=4pt, 
    boxrule=0.8pt,             
    width=\linewidth
}

\section{Prompt Details and Examples}
\label{app:prompt_details}
\rev{To make the evaluation and construction procedures auditable, the following figures reproduce the active compact-v2 judge prompt and representative prompt templates used for benchmark construction, quality screening, and review processing.}

\begin{figure}[!htbp]
    \begin{promptbox}{Holistic VLM Judge Prompt (w. GT / w.o. GT)}
        \textbf{System message:} You are a conservative medical multimodal evaluator. Evaluate the candidate against the supplied evidence. Give the model's own raw ordinal judgments; do not calculate, map, or cap any score. Every required response field must be specific, non-empty, and grounded in the supplied image(s), instruction, and candidate content.

        \vspace{0.5em}
        \textbf{Reference-aware branch (w. GT):} For image tasks, Picture~1 is the reference image and Picture~2 is the candidate/output image. For VQA, Picture~1 is the input image and the reference text/answer is supplied separately. The candidate is judged against both the clinical evidence and the available reference.

        \vspace{0.5em}
        \textbf{Reference-free branch (w.o. GT):} For image tasks, Picture~1 is the input image and Picture~2 is the candidate/output image. For VQA, only the input image, question, and candidate answer are supplied. No reference image or answer is revealed.

        \vspace{0.5em}
        \textbf{Analyze--judge procedure:}
        \begin{enumerate}
            \item Observe every supplied picture independently before comparing it with the candidate or reference text.
            \item For image tasks, assess case identity using case-specific contours, orientation, lesion location and shape, devices, annotations, boundaries, and global content.
            \item For VQA, determine whether the candidate directly answers the stated question and is supported by the image; in the reference-aware view, also compare it with the reference answer.
            \item For an image-task score of 4 or 5, name at least two concrete matching visual anchors and the visible requested change. For a VQA score of 4 or 5, cite concrete image evidence supporting the answer.
            \item Score exactly anatomical accuracy, clinical-finding accuracy, instruction compliance, cross-modal consistency, and hallucination/omission control. Then provide an independent raw holistic score; do not compute it as an arithmetic mean.
        \end{enumerate}

        \textbf{Response requirements:} Every schema field must be filled. Observations and the overall rationale are limited to 18 words, dimension evidence to 10 words, and image-pair comparison evidence to 14 words. Generic statements such as ``consistent'' or ``successfully edited'' are not accepted as evidence, and the full question, reference, or candidate answer must not be restated.

        \textbf{Severity anchors:} 5 = fully supported with no identifiable clinically or task-relevant error; 4 = substantively correct with only a minor non-critical issue; 3 = partially correct or uncertain; 2 = major error, preservation failure, or incomplete target; 1 = unrelated/incorrect content, critical clinical error, or unusable output.

        \vspace{0.5em}
        \textbf{Case payload:} FORMAT: \{format\}; VIEW: \{reference-aware or reference-free\}; INSTRUCTION: \{instruction\}; REFERENCE TEXT/ANSWER: \{reference, when available\}; CANDIDATE TEXT/ANSWER: \{candidate\}. The model is instructed to return only the object required by the response contract.
    \end{promptbox}
    \centering
    \caption{\rev{Compact-v2 holistic-judge prompt reproduced to document the scoring procedure used for both reference-aware (w. GT) and reference-free (w.o. GT) evaluation. It reports the evidence branches, analyze--judge procedure, response requirements, and the current 1--5 severity anchors. Format-specific clauses specialize the shared AA, CFA, IC, CMC, and HOC dimensions without changing the reported metric family.}}
    \label{app:compact_v2_prompt}
    \label{app:VLM_Holistic_Prompt}
    \label{app:VLM_Holistic_Prompt_WO_GT}
\end{figure}

\begin{figure}[!htbp]
    \centering
    \begin{promptbox}{Style Transfer Subtask Example Prompt for Image Editing}
        Convert this CT scan into a professional 3D volume-rendering visualization using advanced volumetric reconstruction techniques.

        \vspace{0.4em}
        \textbf{Technical Implementation:} Apply sophisticated ray-casting algorithms with appropriate opacity transfer functions to create depth-accurate three-dimensional representation while maintaining anatomical fidelity and diagnostic clarity.

        \vspace{0.4em}
        \textbf{Medical Accuracy Requirements:} Ensure all liver anatomical relationships, proportional scaling, and diagnostic features are preserved with clinical validity. Clearly delineate tissue boundaries, structural interfaces, and pathological regions without introducing artifacts.

        \vspace{0.4em}
        \textbf{Quality Standards:} Achieve spatial resolution preservation above 90\%, depth perception accuracy with proper gradient shading, and tissue boundary clarity meeting professional medical imaging standards.

        \vspace{0.4em}
        \textbf{Contextual Considerations:} Hepatocellular carcinoma with portal vein invasion.

        \vspace{0.4em}
        \textbf{Output Validation:} The resulting image must pass medical imaging quality assessment (e.g., PSNR/SSIM validation), maintain full diagnostic accuracy, demonstrate successful style transformation, and preserve all clinically relevant anatomical and pathological information.
    \end{promptbox}
    \vspace{-0.9em}
    \caption{\rev{Style-transfer example reproduced to document how image-editing instructions specify the requested transformation, medical-accuracy constraints, quality standards, clinical context, and output validation.}}
    \label{app:style_transfer_detailed_prompt}
\end{figure}

\newpage

\FloatBarrier
\section{Benchmark-Construction Prompt Examples}
\label{app:construction_prompts}

\rev{The following figures document prompts used for benchmark synthesis, automated quality screening, and postprocessing. Where scored review is required, the prompts use the shared 1--5 AA, CFA, IC, CMC, and HOC dimensions together with an independent Overall score.}

\begin{figure}[!htbp]
    \centering
    \begin{promptbox}{Question Answer Subtask Example Prompt for VQA}
        Analyze the provided medical image and generate a detailed diagnostic explanation, including:
        \begin{itemize}
            \item The most likely diagnosis
            \item Key imaging features supporting the diagnosis
            \item Differential considerations
            \item Clinical implications
        \end{itemize}
        \vspace{0.5em}
        [Answer in 4-6 complete sentences using professional radiological terminology]
    \end{promptbox}
    \vspace{-0.9em}
    \caption{\rev{Question-answer example reproduced to document how VQA construction prompts request a diagnosis, supporting image features, differential considerations, and clinical implications.}}
    \label{app:Question Answer}
\end{figure}

\begin{figure}[!htbp]
    \centering
    \begin{promptbox}{Anatomical Annotation Subtask Example Prompt}
        You are a medical imaging expert. You will receive two images (Input, Ground Truth). Please: (1) infer modality and view; (2) infer the specific anatomical entity/region involved from the images; (3) generate a task instruction by inserting the inferred entity into an instruction template; (4) generate a concise answer describing the Ground Truth image; (5) summarize key visual differences.

        \textbf{Instruction Generation Requirement:}
        Create an instruction that tasks the model with adding anatomical annotations to this medical image, highlighting key structures or pathological findings.

        Return strictly JSON:
        \begin{verbatim}
{
  "instruction": {"content": "...", "explanation": "why the instruction matches 
  the Input image"},
  "answer": {"content": "concise, one-line description, like 'angiomyolipoma'",
  "explanation": "why the answer matches the Ground Truth image"},
  "visual_differences": {"location": "...", "size": "...", "intensity": "...", 
  "texture": "..."}
}
        \end{verbatim}
    \end{promptbox}
    \vspace{-0.9em}
    \caption{\rev{Anatomical-annotation example reproduced to document how paired images are converted into a grounded instruction, a concise reference answer, and structured visual differences.}}
    \label{app:Anatomical Annotation}
\end{figure}

\begin{figure}[!htbp]
    \centering
    \begin{promptbox}{Artifact Removal Subtask Example Prompt}
        You are a medical imaging expert. You will receive two images (Input, Ground Truth). Please generate aligned instruction-answer text pairs based on paired input and ground truth images.

        \textbf{Instruction Generation Requirement:}
        Create an instruction that tasks the model with removing imaging artifacts from this medical scan while preserving all anatomical structures.

        \textbf{Text Pair Generation Requirements:}
        1. \textbf{Semantic Alignment}: The instruction must be grounded in the \textit{Input} image, and the answer must accurately describe the \textit{Ground Truth} image.
        2. \textbf{Factual Accuracy}: Both the instruction and the answer must be factually correct---especially critical in medical or clinical contexts.

        Return JSON format with instruction and answer fields.
        \vspace{0.5em}
        \textbf{Dataset:} DRVD, SMMILE
    \end{promptbox}
    \vspace{-0.9em}
    \caption{\rev{Artifact-removal example reproduced to document how paired images are converted into an editing instruction and reference answer while preserving anatomy and factual accuracy.}}
    \label{app:Artifact Removal}
\end{figure}

\begin{figure}[!htbp]
    \centering
    \begin{promptbox}{Contrast Enhancement Subtask Example Prompt}
        You are a medical imaging expert. Generate instruction-answer pairs for medical image processing.

        \textbf{Instruction Generation Requirement:}
        Create an instruction that tasks the model with enhancing the contrast and visibility of anatomical structures in this medical image.

        \textbf{Requirements:}
        - Instruction must be medically accurate and appropriate
        - Answer should describe the enhanced image clearly
        - Preserve all anatomical relationships and diagnostic features

        Return JSON with instruction and answer content.
        \vspace{0.5em}
        \textbf{Dataset:} DRVD, SMMILE
    \end{promptbox}
    \vspace{-0.9em}
    \caption{\rev{Contrast-enhancement example reproduced to document how construction prompts request improved visibility while preserving anatomical relationships and diagnostic features.}}
    \label{app:Contrast Enhancement}
\end{figure}

\begin{figure}[!htbp]
    \centering
    \begin{promptbox}{Noise Reconstruction Subtask Example Prompt}
        You are a medical imaging expert. Analyze paired medical images and generate instruction-answer pairs.

        \textbf{Instruction Generation Requirement:}
        Create an instruction that tasks the model with reducing noise and reconstructing the image quality while maintaining anatomical accuracy.

        \textbf{Key Requirements:}
        - Maintain diagnostic accuracy during noise reduction
        - Preserve tissue boundaries and structural details
        - Ensure clinical validity of the processed image

        Return structured JSON with instruction, answer, and visual differences.
        \vspace{0.5em}
        \textbf{Dataset:} DRVD, SMMILE
    \end{promptbox}
    \vspace{-0.9em}
    \caption{\rev{Noise-reconstruction example reproduced to document how construction prompts request noise reduction while preserving tissue boundaries, structural detail, and diagnostic content.}}
    \label{app:Noise Reconstruction}
\end{figure}

\begin{figure}[!htbp]
    \centering
    \begin{promptbox}{Resolution Edit Subtask Example Prompt}
        You are a medical imaging expert. Generate aligned text pairs for medical image enhancement tasks.

        \textbf{Instruction Generation Requirement:}
        Create an instruction that tasks the model with improving the resolution and clarity of this medical image.

        \textbf{Medical Accuracy Requirements:}
        - Ensure all anatomical relationships are preserved
        - Maintain proportional scaling and diagnostic features
        - Preserve tissue boundaries and structural interfaces

        Return JSON format with detailed instruction and concise answer.
        \vspace{0.5em}
        \textbf{Dataset:} DRVD, SMMILE
    \end{promptbox}
    \vspace{-0.9em}
    \caption{\rev{Resolution-edit example reproduced to document how construction prompts request improved clarity while preserving scale, anatomy, tissue boundaries, and diagnostic features.}}
    \label{app:Resolution Edit}
\end{figure}

\begin{figure}[!htbp]
    \centering
    \begin{promptbox}{Style Transfer Subtask Example Prompt}
        You are a medical imaging expert. Generate instruction-answer pairs for medical image style transformation.

        \textbf{Instruction Generation Requirement:}
        Create an instruction that tasks the model with applying a different imaging style or modality appearance to this medical image while preserving anatomical accuracy.

        \textbf{Quality Standards:}
        - Achieve spatial resolution preservation above 90\%
        - Maintain depth perception accuracy with proper gradient shading
        - Preserve all clinically relevant anatomical and pathological information

        Return JSON with instruction content, answer content, and explanations.
        \vspace{0.5em}
        \textbf{Dataset:} DRVD, SMMILE
    \end{promptbox}
    \vspace{-0.9em}
    \caption{\rev{Style-transfer construction example reproduced to document how prompts request an appearance transformation while retaining spatial resolution, depth cues, anatomy, and pathology.}}
    \label{app:Style Transfer}
\end{figure}

\begin{figure}[!htbp]
    \centering
    \begin{promptbox}{Organ Removal Subtask Example Prompt}
        You are a medical imaging expert. Enrich the instruction based on annotation information and sample images.

        \textbf{Task:} Organ Removal
        \textbf{Annotation Information:}
        - Organ: \{organ\} (e.g., liver, kidney, heart)
        - Removal Method: \{removal\_method\} (e.g., noise\_fill, gaussian\_blur, black\_fill)
        - Modality: \{modality\} (e.g., CT, MRI, Ultrasound)

        \textbf{Instruction Templates (by method):}
        - noise\_fill: "Remove the \{organ\} and fill the region with noise in this \{modality\} image"
        - gaussian\_blur: "Remove the \{organ\} and apply Gaussian blur to the region in this \{modality\} image"
        - black\_fill: "Remove the \{organ\} and fill the region with black in this \{modality\} image"

        \textbf{Complexity Levels:}
        - Basic: Simple instruction with organ and modality
        - Detailed: Add quality descriptors and view information
        - Clinical: Include clinical diagnostic terminology

        Generate enriched instruction following the selected template and complexity level.
        \vspace{0.5em}
        \textbf{Dataset:} U2Bench
    \end{promptbox}
    \vspace{-0.9em}
    \caption{\rev{Organ-removal example reproduced to document how annotation fields, removal methods, modality, and complexity level are combined into enriched construction instructions.}}
    \label{app:Organ Removal}
\end{figure}

\begin{figure}[!htbp]
    \centering
    \begin{promptbox}{Organ Reconstruction Subtask Example Prompt}
        You are a medical imaging expert. Create enriched instructions for organ reconstruction tasks.

        \textbf{Task:} Organ Reconstruction
        \textbf{Annotation Information:}
        - Organ: \{organ\}
        - Reconstruction Method: \{method\} (e.g., inpainting, interpolation)
        - Modality: \{modality\}

        \textbf{Instruction Templates:}
        - inpainting: "Reconstruct the \{organ\} using inpainting in this \{modality\} image"
        - interpolation: "Reconstruct the \{organ\} using interpolation in this \{modality\} image"

        \textbf{Enrichment Strategy:}
        1. Select template based on annotation method
        2. Add modality-specific quality descriptors (detailed level)
        3. Include clinical terminology (clinical level)

        Return enriched instruction with metadata indicating complexity and enrichment method.
        \vspace{0.5em}
        \textbf{Dataset:} U2Bench
    \end{promptbox}
    \vspace{-0.5em}
    \vspace{-0.9em}
    \caption{\rev{Organ-reconstruction example reproduced to document how annotation fields, reconstruction methods, modality, and complexity level are combined into enriched construction instructions.}}
    \label{app:Organ Reconstruction}
\end{figure}

\begin{figure}[!htbp]
    \centering
    \begin{promptbox}{Instruction Edit Subtask Example Prompt}
        You are a medical imaging expert. Generate enriched instructions for image editing tasks.

        \textbf{Task:} Instruction Edit
        \textbf{Annotation Information:}
        - Edit Method: \{method\} (e.g., brighten, contrast, sharpen)
        - Organ: \{organ\}
        - Modality: \{modality\}

        \textbf{Instruction Templates:}
        - brighten: "Brighten the \{organ\} region in this \{modality\} image"
        - contrast: "Enhance the contrast of the \{organ\} in this \{modality\} scan"
        - sharpen: "Sharpen the \{organ\} structure in this \{modality\} image"

        \textbf{Smart Enrichment:}
        - Use annotation-based template selection
        - Apply complexity-based enhancement (basic/detailed/clinical)
        - Preserve medical accuracy and clinical validity

        Generate instruction following smart enrichment pipeline.
        \vspace{0.5em}
        \textbf{Dataset:} Custom
    \end{promptbox}
    \vspace{-0.5em}
    \vspace{-0.9em}
    \caption{\rev{Instruction-edit example reproduced to document how edit method, organ, modality, and complexity level guide the enrichment of image-editing instructions.}}
    \label{app:Instruction Edit}
\end{figure}

\begin{figure}[!htbp]
    \centering
    \begin{promptbox}{Dye Transfer Subtask Example Prompt}
        You are a medical imaging expert. Create enriched instructions for histological staining transfer.

        \textbf{Task:} Dye Transfer
        \textbf{Annotation Information:}
        - Source Staining: \{source\_staining\} (e.g., H\&E)
        - Target Staining: \{target\_staining\} (e.g., Masson, PAS)
        - Modality: \{modality\}

        \textbf{Instruction Templates:}
        - "Apply computational stain transfer from \{source\_staining\} to \{target\_staining\} staining using AI-based color mapping"
        - "Transform this \{source\_staining\} histology slide to \{target\_staining\} staining appearance using deep learning methods"

        \textbf{Enrichment:}
        - Select template based on staining types
        - Add technical implementation details (detailed level)
        - Include tissue architecture preservation requirements (clinical level)

        Return enriched instruction with staining transfer specifications.
        \vspace{0.5em}
        \textbf{Dataset:} DRVD
    \end{promptbox}
    \vspace{-0.9em}
    
    \caption{\rev{Dye-transfer example reproduced to document how source and target stains, technical detail, and tissue-preservation requirements guide enriched histology instructions.}}
    \label{app:Dye Transfer}
\end{figure}

\begin{figure}[!htbp]
    \centering
    \begin{promptbox}{Quality Validation Prompt}
        You are a medical imaging expert. Review whether this candidate is suitable for medical-AI training.

        \textbf{Inputs:} Instruction: \{instruction\}; Answer: \{answer\}; Modality: \{modality\}; Task: \{task\_type\}.

        \textbf{Score each item from 1 (unacceptable) to 5 (fully supported):} AA---anatomy, site, and laterality; CFA---clinical findings and details; IC---instruction fulfillment; CMC---image--text consistency; HOC---absence of unsupported claims and clinically important omissions.

        Give one short rationale for each score. Then set \texttt{retain} to \texttt{true} only when all five scores are at least 4 and no material error or unsupported clinical claim is present; otherwise set it to \texttt{false}.

        \textbf{Return JSON only:}
        \begin{verbatim}
{
  "AA": <1-5>, "CFA": <1-5>, "IC": <1-5>,
  "CMC": <1-5>, "HOC": <1-5>,
  "rationale": {"AA": "...", "CFA": "...", "IC": "...",
                "CMC": "...", "HOC": "..."},
  "retain": <true|false>,
  "decision_reason": "<brief reason>"
}
        \end{verbatim}

        \vspace{0.5em}
        \textbf{Dataset:} All processed datasets
    \end{promptbox}
    \vspace{-0.9em}
    \caption{\rev{Quality-validation prompt reproduced to document five-dimensional scoring, concise supporting rationales, and a rule-based Boolean retention decision for candidate instruction--answer pairs.}}
    \label{app:Quality Validation}
    
\end{figure}

\begin{figure}[!htbp]
    \centering
    \begin{promptbox}{Percentage Filtering Prompt}
        You are a medical imaging expert. Filter JSONL file samples by removing the lowest-scoring percentage.

        \textbf{Task:} Filter Lowest Percentage
        \textbf{Input:}
        - JSONL file with current clinical evaluation scores
        - Filter percentage: \{filter\_percent\} (default: 20\%)

        \textbf{Process:}
        1. Check whether Overall, AA, CFA, IC, CMC, and HOC fields exist in each item
        2. If missing, perform the current VLM clinical evaluation for all samples
        3. Rank items using their independent Overall score, supported by the five recorded dimensions
        4. Sort items by Overall score (descending)
        5. Remove lowest \{filter\_percent\}\% of samples
        6. Keep top (1-\{filter\_percent\})\% samples

        \textbf{Output:}
        - Filtered JSONL file with only high-quality samples
        - Statistics: max/min/avg/median scores
        - Removed samples score range

        Filter the dataset and save filtered results.
        \vspace{0.5em}
        \textbf{Dataset:} All validated datasets
    \end{promptbox}
    \vspace{-0.9em}
    \caption{\rev{Percentage-filtering prompt reproduced to document how low-scoring samples are ranked and removed after clinical quality screening.}}
    \label{app:Percentage Filtering}
\end{figure}

\begin{figure}[!htbp]
    \centering
    \begin{promptbox}{Review Processing Prompt}
        You are a medical imaging expert. Process VLM review tasks for MMGeneration datasets.

        \textbf{Task:} VLM Review Processing
        \textbf{Subtasks:}
        - 2d-3d-reconstruction
        - 3d-2d-projection
        - dye-transfer
        - instruction-edit
        - organ-reconstruction
        - organ-removal

        \textbf{Quality Thresholds:}
        - overall\_holistic\_score >= 4
        - anatomical\_accuracy >= 4
        - clinical\_finding\_accuracy >= 4
        - instruction\_compliance >= 4
        - cross\_modal\_consistency >= 4
        - hallucination\_omission\_control >= 4

        \textbf{Process:}
        1. Load the current clinical evaluation report from JSON file
        2. Filter valid samples based on quality thresholds
        3. Improve instructions for valid samples:
           - Use task-specific templates
           - Add modality-specific descriptors
           - Include clinical terminology
        4. Improve answers for valid samples
        5. Save improved samples, valid samples, and invalid samples

        \textbf{Output Files:}
        - \{filename\}\_improved.jsonl (all samples with improvements)
        - \{filename\}\_valid.jsonl (only valid samples)
        - \{filename\}\_invalid.jsonl (only invalid samples)

        Process the review tasks and generate improved datasets.
        \vspace{0.5em}
        \textbf{Dataset:} MMGeneration (M3D, U2Bench, MedFrameQA, DRVD)
    \end{promptbox}
    \vspace{-0.9em}
    \caption{\rev{Review-processing prompt reproduced to document how clinically valid multimodal-generation samples are selected, refined, and partitioned into output files.}}
    \label{app:Review Processing}
\end{figure}

\FloatBarrier
\revisionoff
\fi
\color{black}


\end{document}